\documentclass[letterpaper,journal]{IEEEtran}
\usepackage{amsmath,amssymb}
\usepackage{graphicx}
\usepackage{booktabs,multirow}
\usepackage{cite}
\usepackage{url}
\usepackage{stfloats}

\begin{document}
\raggedbottom

\title{IRadiDINO: Radiometric Variation-Aware Domain Generalization for Aerial Thermal Infrared Vehicle Detection}


\author{Mengjie Cao, Bingwei Hui, Zhongzhen Sun, Zhenfeng Shao, and Ronghui Zhan%
\thanks{This work was supported by the Innovation Research Foundation of National University of Defense Technology. \textit{(Corresponding author: Bingwei Hui.)}}%
\thanks{Mengjie Cao, Bingwei Hui, Zhongzhen Sun, and Ronghui Zhan are with the ATR Key Laboratory, College of Electronic Science and Technology, National University of Defense Technology, Changsha 410073, China (e-mail: caomengjie@nudt.edu.cn; huibingwei07@nudt.edu.cn; sunzhongzhen14@nudt.edu.cn; zhanrh@nudt.edu.cn).}%
\thanks{Zhenfeng Shao is with the State Key Laboratory of Information Engineering in Surveying, Mapping and Remote Sensing, Wuhan University, Wuhan 430079, China (e-mail: shaozhenfeng@whu.edu.cn).}}

\maketitle

\begin{abstract}

Cross-dataset UAV-borne thermal infrared vehicle detection is susceptible to shifts in radiometric statistics and interference from complex backgrounds. The former changes the radiometric responses of vehicles and their surroundings, whereas the latter can produce local feature responses resembling those of vehicles. Both factors degrade vehicle detection and localization. Although vision foundation models provide generic semantic representations with transfer potential, adaptation under source-domain supervision may still rely on specific radiometric appearances. To address these issues, we propose IRadiDINO, a DINOv3-based single-source domain generalization framework that improves cross-scene feature representations through adaptation to radiometric variations and local spatial discrimination. First, radiometric intervention for low-rank semantic adaptation (RI-LoRA) uses symmetric attenuation and amplification of low-frequency amplitudes to construct multiple radiometric states. This exposes shared low-rank parameters to detection supervision under varying radiometric conditions and reduces reliance on source-specific radiometric statistics. Second, the semantic-anchored ordinal relation spatial prior (SAOR) jointly encodes signed radiometric ordinal relations normalized by local scale, feature differences, and central semantic anchors to better distinguish vehicle structures from complex backgrounds. Experiments use four public infrared vehicle datasets (DroneVehicle, ESCVehicle, M3OT, and HIT-UAV) and an additional dataset collected by our team, VIVID. Its dedicated evaluation set contains 2,000 images and extends the evaluation to scenes with complex land-cover backgrounds and small vehicles. In bidirectional cross-domain bounding-box detection with DroneVehicle and ESCVehicle as the respective source domains, the DINOv3 variant achieves mean out-of-domain (OOD) AP50/F1 scores of 59.00\%/66.21\% and 72.35\%/78.94\%, respectively. With DroneVehicle as the source, it achieves a Point-AP of 59.89\% for center-point detection on ESCVehicle, exceeding the strongest comparison method by 25.59 percentage points. Under the same DINOv2-L/14 backbone, our method also outperforms Bridge in both mean OOD AP50 and F1, indicating that the gains cannot be attributed solely to the pretrained backbone upgrade. The results show that the framework accommodates radiometric variations while improving local spatial discrimination, providing an effective approach to cross-dataset UAV-borne thermal infrared vehicle detection under complex radiometric conditions and background interference.

\end{abstract}


\section{Introduction}

\IEEEPARstart{U}{AV}-borne thermal infrared remote sensing captures thermal radiation from ground vehicles and their surroundings, enabling observation at night and under low illumination for applications such as traffic monitoring, road safety patrols, surveillance of key areas, and emergency search after disasters~\cite{ref1,ref2,ref3,ref4}. However, diffraction, ambient thermal radiation noise, and atmospheric transmission in long-wave infrared imaging affect radiometric responses and spatial detail~\cite{ref5}. Variations in the thermal state and composition of scene backgrounds can also alter the local radiometric appearance of vehicles, complicating recognition and localization in complex scenes~\cite{ref6,ref7}. In practice, detectors are deployed across regions, environmental conditions, and observation platforms. Differences in object thermal states, surface radiative properties, and sensor responses change the appearance of vehicles and surrounding terrain, limiting the direct transfer of features learned from a particular data distribution to unseen scenes. Given the cost of recollecting and annotating infrared data for new deployments~\cite{ref8} and the effect of annotation quality on detection performance~\cite{ref9}, reliable cross-scene vehicle detection using only existing source-domain supervision is important~\cite{ref10,ref11}.

Studies of domain-generalized object detection mainly use visible-light images of natural scenes to examine instance-level recognition and bounding-box localization across scenes~\cite{ref10,ref12}. Research on infrared small and dim target detection and its cross-dataset generalization instead emphasizes enhancing weak responses and separating targets from complex backgrounds, often through pixel-level segmentation~\cite{ref13,ref14}. UAV-borne thermal infrared vehicle detection is an instance-level task that must accommodate scene-dependent imaging variations while preserving vehicle discriminability and bounding-box localization accuracy in complex backgrounds.

Large-scale vision foundation models provide a new basis for feature learning. Pretraining on large and diverse image collections yields rich generic representations that can transfer to downstream vision tasks~\cite{ref15,ref16}. However, such transferability does not guarantee that these models directly accommodate thermal infrared imaging or cross-scene detection. Thermal infrared images typically contain limited texture and low local contrast, and their radiometric representations are also affected by scene temperature and task requirements~\cite{ref17}. When adapted using labeled source-domain data, the model may still use source-specific radiometric statistics as discriminative cues, making it difficult to handle changes in vehicle appearance and vehicle--background thermal contrast across scenes. Furthermore, vehicles often appear small in UAV imagery, and structured backgrounds such as roads, buildings, and vegetation can interfere with the extraction and discrimination of local vehicle features~\cite{ref18}. A central challenge is therefore to exploit pretrained visual knowledge while reducing the dependence of semantic adaptation on source-specific radiometric statistics and improving the modeling of vehicle-related local spatial relations.

We propose a vision-foundation-model-based framework for domain-generalized infrared vehicle detection. It uses DINOv3 as the pretrained backbone, an existing feature transformation structure~\cite{ref10} to construct multiscale representations, and a two-stage detector to predict vehicle bounding boxes. To address the dependence on source-domain radiometric statistics that may arise during semantic adaptation under single-source supervision, we propose radiometric intervention for low-rank semantic adaptation (RI-LoRA), which allows the low-rank parameters to learn detection features under varying radiometric conditions. We also propose the semantic-anchored ordinal relation spatial prior (SAOR), which incorporates local relative radiometric variations and pretrained semantic information into spatial relation modeling to provide fine-grained structural features with a semantic reference for vehicle recognition and localization in complex scenes. We evaluate the method through cross-dataset detection experiments on multiple infrared vehicle datasets, comparisons with representative methods, and ablation studies. The main contributions are as follows.

1) We present the first systematic study of cross-domain generalization for vehicle detection in infrared remote sensing. Under a strict single-source training protocol with a unified vehicle category and evaluation protocol, we evaluate representative methods on bounding-box detection and center-point localization through bidirectional transfer and additional unseen-domain tests. We also introduce VIVID, collected by our team, to assess generalization in scenes with complex backgrounds and small vehicles.

2) We propose RI-LoRA. Symmetric attenuation and amplification of low-frequency amplitudes diversify source-domain radiometric states, exposing the shared low-rank parameters to detection supervision under varying radiometric conditions and reducing the dependence of semantic adaptation on source-specific radiometric statistics.

3) We propose SAOR. Signed ordinal relations normalized by local scale represent relative radiometric variations, and pretrained semantic anchors are used to jointly encode local feature differences and radiometric ordinal relations. The resulting fine-grained spatial representation incorporates both local radiometric variations and discriminative semantic information, providing a spatial prior for vehicle recognition and localization in complex backgrounds.

\section{Related Work}

\subsection{Infrared Vehicle Detection}

Infrared vehicle detection primarily addresses target enhancement, fine-grained feature extraction, and multiscale fusion under limited imaging information. Yu et al.~\cite{ref19} proposed RYOLOv5\_D, which highlights vehicles through adaptive histogram enhancement (AAHE) and combines a convolutional polarized self-attention block (CPSAB) with deformable convolution to improve target responses and adaptation to target shapes in complex backgrounds. Zhang et al.~\cite{ref20} proposed NOC-YOLO, which improves convolution, the backbone, and upsampling to enhance fine-grained feature extraction and multiscale information exchange. Wang et al.~\cite{ref21} proposed LFIR-YOLO, which incorporates a dilation-wise residual (DWR) module into the backbone, fuses features from multiple levels through content-guided attention (CGA), and improves feature representations through detail-enhanced convolution (DEConv). These methods improve infrared vehicle detection primarily through target response enhancement and feature extraction.

Other studies have investigated local radiometric patterns and spatial structure. Zhang et al.~\cite{ref22} proposed OSIV-Net, which uses anchor-adaptive convolution (A\textsuperscript{2}Conv), combining differential and dynamic convolutions, to extract relatively stable vehicle edge and corner patterns as well as variable local radiometric patterns. Its Magic Cube Module (MCM) aggregates global information in the three-dimensional feature space. Pei et al.~\cite{ref23} proposed Fs2PA, which introduces a geometric boundary prior through gradient-guided attention (GIA) and uses a scale-aware shared detection head (SAS-Head) to impose cross-scale constraints between deep P\textsubscript{5} semantics and shallow P\textsubscript{2} features. These studies broaden the approaches to local pattern extraction and spatial structure modeling.

These two approaches enhance vehicle spatial representations by extracting local variation patterns and by linking geometric and semantic information, respectively. Their modeling focuses on local differential responses, context aggregation, and cross-scale feature transfer. A unified, explicit representation of directional radiometric variations between a central region and its neighborhood and of their relationship to the central semantics remains lacking.

\subsection{Distribution Diversification for Domain Generalization}

Domain-generalized object detection typically broadens the source training distribution to improve performance in unseen scenes. Appearance perturbations must also preserve object structure and localization information. Lee et al.~\cite{ref24} proposed OA-DG, which generates diverse samples using multilevel image transformations and object-aware mixing and mitigates semantic loss by learning domain-invariant object and background representations. Danish et al.~\cite{ref11} proposed DivAlign, which combines multiple image augmentations to expand the source distribution and aligns class predictions and bounding-box localization across views. Lu et al.~\cite{ref25} proposed a statistics-driven adaptive piecewise mapping method (SDAPM) that adjusts mapping parameters according to image statistics and simulates different weather conditions with a dynamic heterogeneous domain perturbation generator, reducing foreground detail loss during distribution expansion. These studies highlight the need to diversify appearance while retaining object information.

Because appearance changes are often accompanied by changes in spatial-frequency components and their statistics, frequency-domain transformations have also been used to diversify training samples. Xu et al.~\cite{ref26} proposed FACT, which mixes Fourier amplitude spectra from different images while preserving the original phase and uses co-teacher regularization to enforce prediction consistency before and after augmentation. Zhang et al.~\cite{ref27} proposed frequency-based pseudo-domain generation (FPG), which exploits frequency-domain style information to construct pseudo-domains and combines rotation prediction with semantic consistency learning to improve cross-domain detection representations. Xu et al.~\cite{ref28} proposed PhysAug, which combines a physical imaging model with frequency-domain perturbations to simulate distribution shifts under nonideal imaging conditions. Wang et al.~\cite{ref29} proposed adversarial spectral perturbation (AdvSP), which introduces learnable perturbations to different spectral components and uses adversarial optimization to generate training samples with varied appearances.

Cross-scene variations in thermal infrared imagery differ from general changes in color, texture, or weather. They also depend on the environmental thermal state, the radiative properties of vehicles and backgrounds, and sensor responses, which alter image radiometric statistics and relative target--background thermal contrast. Existing frequency-domain methods use amplitude mixing, low-frequency adjustment, or perturbations of multiple spectral components, but are primarily designed for general visual domain generalization, natural image style variation, or simulation of nonideal optical imaging. These objectives only partially capture radiometric state changes in thermal infrared scenes. Furthermore, the range and strength of spectral perturbations can affect local vehicle structures and target--background relations. Expanding the distribution for infrared vehicle detection therefore requires controllable frequency-domain interventions informed by its radiometric statistics while preserving vehicle structure and discriminative information under different interventions.

\subsection{Generalization and Infrared Transfer of VFMs}

Large-scale self-supervised pretraining provides vision foundation models with generic representations that can transfer across tasks. DINOv2, proposed by Oquab et al.~\cite{ref16}, uses large-scale self-supervised learning to obtain rich semantic and dense features. Sim\'{e}oni et al.~\cite{ref30} subsequently proposed DINOv3, which scales up the pretraining data and model and introduces Gram Anchoring to improve dense feature quality. However, generic pretrained representations do not automatically accommodate distribution shifts in unseen scenes. He et al.~\cite{ref31} proposed the Generalized Diffusion Detector (GDD), which transfers generative pretraining knowledge to object detection using intermediate denoising features and feature- and object-level alignment. Hong et al.~\cite{ref32} proposed Bridge, which learns low-rank bases for causal front-door adjustment to mitigate confounding effects such as illumination, object co-occurrence, and image style. Zhang et al.~\cite{ref33} proposed VFM\textsuperscript{4}SDG, which uses a frozen vision foundation model to construct a cross-domain stability prior, preserves object--background and inter-instance feature relations through relational distillation, and enhances detection queries using class semantic prototypes and global object context. These studies extend domain generalization through generative feature transfer, causal feature calibration, and cross-domain relational priors, respectively. Their relation modeling primarily concerns distillation of cross-domain stable relations and enhancement of detection queries. Directional relative radiometric variations between a center and its neighbors in thermal infrared images and their locally conditioned association with central semantics remain unaddressed.

To address the modality gap between natural-image pretraining and infrared imaging, several studies have developed infrared-specific vision foundation models. Liu et al.~\cite{ref34} proposed InfMAE, constructed Inf30 with 305,241 infrared images, and learned infrared features using information-aware masking and a multiscale encoder. Xing et al.~\cite{ref35} constructed Inf-590K with 590,700 images and proposed token entropy-based masking. They incorporated frequency-domain information through a dual-domain guidance module (DDG) and adaptive frequency-domain modulation (AFDM) to improve infrared feature learning and suppress nonuniform background noise. These studies incorporate the information distribution and radiometric characteristics of infrared imaging into foundation model pretraining, highlighting the importance of modality-specific imaging statistics and information structure. Infrared-specific models primarily use large-scale self-supervised infrared pretraining, a different approach from adapting existing generic visual knowledge to downstream tasks. In particular, AFDM in DuGI-MAE emphasizes adaptive frequency-domain filtering and nonuniform noise suppression, whereas our approach constructs radiometric states through low-frequency amplitude interventions to optimize shared low-rank adaptation parameters.

Another approach transfers general-purpose vision foundation models to infrared tasks. Zhang et al.~\cite{ref36} proposed IRSAM, which uses a structure enhancement module and a granularity-aware decoder to improve feature extraction for infrared small and dim targets. Fan et al.~\cite{ref37} applied RGB-pretrained DINOv2 to thermal infrared depth estimation and used a two-stage training strategy with low-rank and convolutional adapters. Nalcakan et al.~\cite{ref38} proposed SpectraDINO to study transfer of DINOv2 across near-, short-wave, and long-wave infrared imagery using modality-specific inputs, lightweight adapters, and feature distillation. A local neighborhood structure preservation constraint maintains feature relations. For downstream adaptation, ViT-Adapter~\cite{ref39} incorporates spatial priors and multiscale feature interaction to improve dense prediction representations, while LoRA~\cite{ref40} reduces adaptation costs through low-rank parameter updates. These studies demonstrate the feasibility of transferring generic visual knowledge to infrared tasks, primarily small target detection, depth estimation, and cross-spectral feature transfer. Instance-level vehicle detection in complex thermal infrared scenes differs from those tasks. It also requires pretrained semantics to account for variations in radiometric statistics across scenes and for the association between local vehicle radiometric relations and semantic information.

\section{Method}

\subsection{Problem Formulation}

We study the generalization of UAV-borne infrared vehicle detection to unseen scenes under distribution shift. During training, the model is supervised using a single labeled infrared vehicle dataset. At test time, the trained detector is applied directly to other datasets to evaluate vehicle detection under different imaging conditions and scene distributions. The test datasets are excluded from training and parameter adaptation.

Given the training dataset
\begin{equation}
D_s=\{(x_i^s,y_i^s)_{i=1}^{N_s}\},
\label{eq:1}
\end{equation}
where $x_i^s$ denotes a source-domain infrared image and $y_i^s$ denotes its detection annotations, comprising the bounding box and class label of each vehicle instance. We merge all vehicle types into a single vehicle class. Let $f_\theta$ denote the detector to be trained. The baseline source-domain optimization objective is
\begin{equation}
\theta^\ast=\arg\min_\theta \mathbb{E}_{(x,y)\sim P_s}\bigl[L_{\mathrm{det}}(f_\theta(x),y)\bigr],
\label{eq:2}
\end{equation}
where $P_s$ denotes the source-domain data distribution and $L_{\mathrm{det}}$ denotes the object detection loss, which covers region proposal generation, object classification, and bounding-box regression. At test time, the trained detector is applied directly to an unseen target domain $D_t$. The target distribution $P_t$ may differ from the source distribution $P_s$, and target images, labels, and statistics are excluded from training and parameter adjustment. The objective is to improve detection on unseen infrared data distributions under single-source supervision, beyond optimizing source-domain performance alone.

\subsection{Overall Framework}

\begin{figure*}[!tb]
\centering
\includegraphics[width=\textwidth,height=0.48\textheight,keepaspectratio]{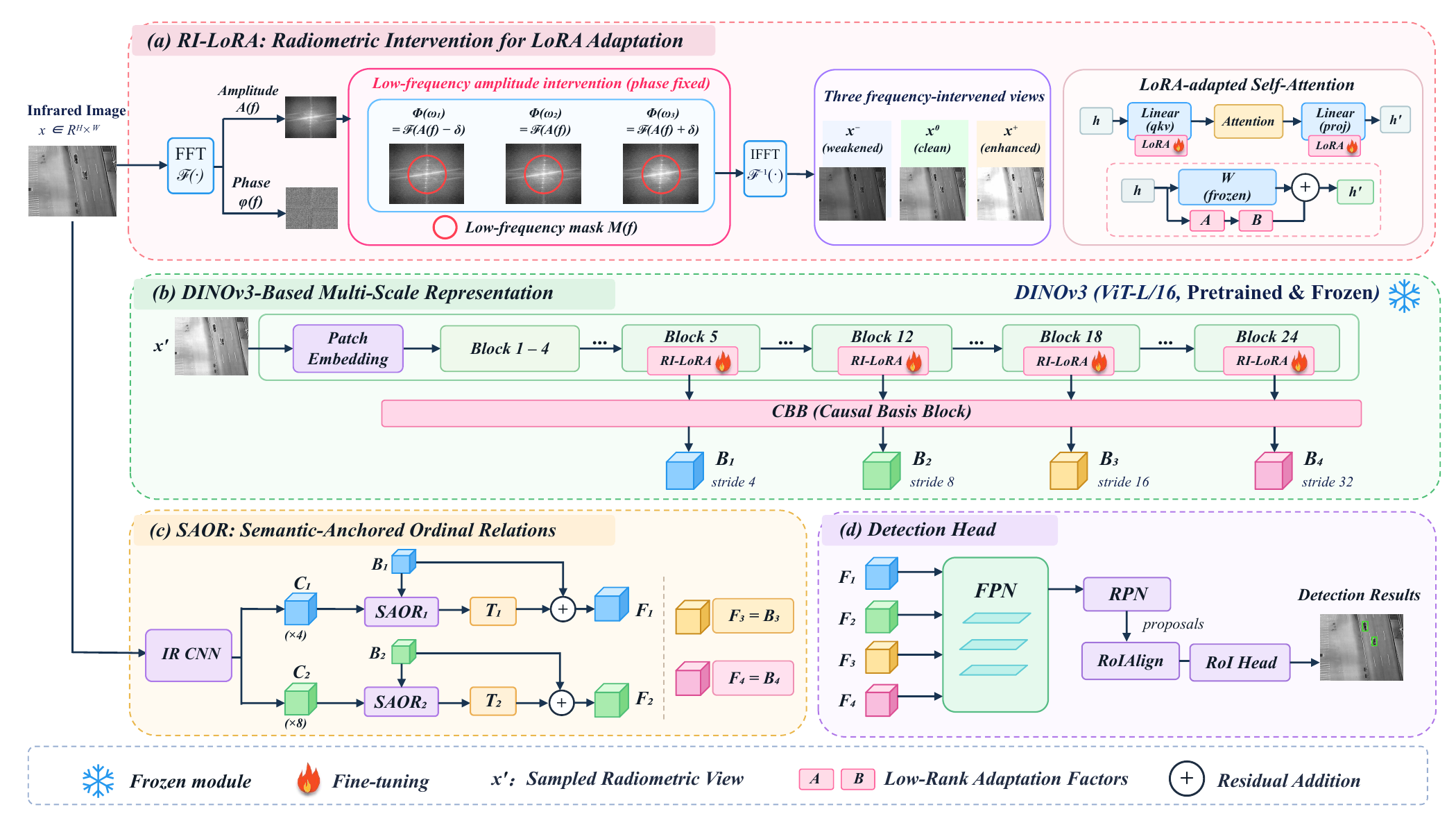}
\caption{Overall framework of the proposed single-source domain-generalized infrared vehicle detector.}
\label{fig:1}
\end{figure*}

As shown in Fig.~\ref{fig:1}, our cross-domain framework for UAV-borne thermal infrared vehicle detection uses a pretrained vision foundation model and comprises a visual semantic branch, an infrared spatial branch, and a detection head. Given an infrared image $x$, the visual semantic branch uses the radiometrically intervened view $x'$ to extract hierarchical semantic features through the pretrained backbone and low-rank adaptation. The infrared spatial branch processes the original image directly to extract local spatial features. SAOR then models local spatial relations between the two feature types to construct spatial priors, which are fused with high-resolution semantic features. The multiscale detection head uses the resulting features to predict vehicles.

The visual semantic branch uses DINOv3 ViT-L/16 pretrained on LVD-1689M. We freeze the original backbone parameters and introduce RI-LoRA in transformer blocks 5, 12, 18, and 24. The normalized spatial tokens output by these blocks are rearranged into two-dimensional feature maps to construct multilevel visual features
\begin{equation}
V=\{V_1,V_2,V_3,V_4\},\quad V_i\in\mathbb{R}^{C\times H_p\times W_p},
\label{eq:3}
\end{equation}
where $C=1024$ is the number of feature channels, and $H_p$ and $W_p$ are the height and width of the spatial token grid, respectively. Because the visual features extracted at different depths have the same spatial resolution, we calibrate them using the causal basis block (CBB) from Bridge~\cite{ref32}. The calibrated features are then mapped to their respective scales to construct four levels of base semantic features
\begin{equation}
\mathcal{B}=\operatorname{CBB}(\mathcal{V})=B_1,B_2,B_3,B_4,
\label{eq:4}
\end{equation}
where the four feature levels have spatial strides of 4, 8, 16, and 32, respectively, supporting subsequent cross-branch feature interaction and multiscale detection.
\begin{equation}
C_1,C_2=\phi_{\mathrm{IR}}(x),
\label{eq:5}
\end{equation}
where $C_1$ and $C_2$ correspond to the feature levels with strides of 4 and 8, respectively. SAOR combines the original image intensity $G$, infrared convolutional features $C_s$, and visual semantic features $B_s$ at the corresponding level to construct two levels of spatial priors
\begin{equation}
T_s=\Psi_{\mathrm{sp}}(B_s,C_s,G),\quad s\in\{1,2\},
\label{eq:6}
\end{equation}
where $\Psi_{\mathrm{sp}}$ denotes local spatial relation modeling and $T_s$ is the spatial prior at the corresponding scale.

During feature fusion, the two infrared spatial priors are fused with the first two base semantic feature levels, respectively. The deeper levels with strides of 16 and 32 retain their semantic representations, yielding four levels of fused multiscale features
\begin{equation}
\mathcal{F}=\operatorname{Fuse}(\mathcal{B},\mathcal{P})=F_1,F_2,F_3,F_4,
\label{eq:7}
\end{equation}
where $\operatorname{Fuse}(\cdot)$ denotes cross-branch feature fusion. Section~\ref{sec:fusion} details the feature mapping and fusion operations. The four fused feature levels are then fed into a feature pyramid network (FPN) to construct multiscale detection representations, and the Faster R-CNN head predicts vehicles. This process is expressed as
\begin{equation}
\widehat{\mathcal{Y}}=\mathcal{H}_{\mathrm{det}}\bigl(\operatorname{FPN}(\mathcal{F})\bigr),
\label{eq:8}
\end{equation}
where $\mathcal{H}_{\mathrm{det}}$ denotes detection using a region proposal network (RPN), RoIAlign, and an RoI head, and $\widehat{\mathcal{Y}}$ denotes the final vehicle class and bounding-box predictions. Sections~\ref{sec:ri-lora}, \ref{sec:saor}, and \ref{sec:fusion} describe radiometric intervention and low-rank adaptation, local spatial relation modeling, and feature fusion and detection optimization, respectively.

\subsection{RI-LoRA: Radiometric Intervention for Low-Rank Semantic Adaptation}\label{sec:ri-lora}

Freezing the pretrained visual backbone preserves its generic visual representations, but the new low-rank parameters are still updated under source-domain detection supervision. Because discriminative vehicle cues and source-specific radiometric appearances co-occur in training samples, parameter updates may be influenced by source-specific imaging conditions and radiometric statistics. To reduce reliance on the radiometric appearance of a single source domain, RI-LoRA introduces controlled low-frequency radiometric variations, thereby exposing the shared low-rank parameters to detection supervision under different radiometric states.

\textbf{Low-Frequency Amplitude Intervention.} Large-scale intensity variations in infrared images are primarily reflected in low-frequency components, whereas local edges and details are more strongly represented by mid- and high-frequency components and phase information. We therefore selectively modulate non-DC low-frequency components of the Fourier amplitude spectrum while preserving the original phase. This alters the overall radiometric appearance while preserving the original spatial structure as much as possible. Given an input infrared image $x$, we first apply the two-dimensional discrete Fourier transform
\begin{equation}
X(x)=\mathcal{F}(x)=A(x)\odot e^{jP(x)},
\label{eq:9}
\end{equation}
where $\mathcal{F}(\cdot)$ denotes the two-dimensional discrete Fourier transform, and $A(x)$ and $P(x)$ denote the amplitude and phase spectra, respectively. We then construct Gaussian weights on the normalized frequency plane centered at zero frequency
\begin{equation}
M_\rho(u,v)=\exp\left[-\frac{f_u^2+f_v^2}{2\rho^2}\right],
\label{eq:10}
\end{equation}
where $(f_u,f_v)$ denotes the normalized frequency coordinates and $\rho$ controls the frequency decay scale of the Gaussian weights. To preserve the global mean image intensity, we set the weight at the DC component to zero
\begin{equation}
M_\rho(0,0)=0.
\label{eq:11}
\end{equation}

The Gaussian weights decay smoothly with distance from zero frequency, concentrating amplitude modulation in the non-DC low-frequency region and mitigating spatial ringing that can arise from hard spectral truncation.

Because large-scale intensity responses can increase or decrease under different radiometric conditions, we construct symmetric positive and negative low-frequency amplitude interventions relative to the original amplitude spectrum. For intervention state $s\in[-1,+1]$ and intensity $\delta$, the modulated amplitude spectrum is defined as
\begin{equation}
A_s(x)=A(x)\odot\bigl(1+s\delta M_\rho\bigr),
\label{eq:12}
\end{equation}
where $s=-1$ and $s=+1$ denote low-frequency amplitude attenuation and amplification, respectively. Because $\delta\in[0.05,0.10]$ and $M_\rho\in[0,1]$, the amplitude scaling factor remains positive, avoiding spectral phase reversal. The original observation is included in training alongside the two amplitude-modulated states, yielding a three-state radiometric intervention. The three states $s\in\{-1,0,+1\}$ are sampled with probabilities $P(s=-1)=0.25$, $P(s=0)=0.5$, and $P(s=+1)=0.25$, respectively. For either nonoriginal state, the intervention intensity is drawn from the uniform distribution $\delta\sim U(0.05,0.10)$. The modulated amplitude spectrum is recombined with the original phase, and the two-dimensional inverse Fourier transform yields the intervened image
\begin{equation}
x_s=\operatorname{Re}\left[F^{-1}\left(A_s(x)\odot e^{jP(x)}\right)\right].
\label{eq:13}
\end{equation}

We denote the intervention by $x_{s,\delta}=T_{s,\delta}(x)$. When $s=0$, we set $\delta=0$, so the intervention is the identity mapping and the original input is used directly without a spectral transform. Because the intervention does not change spatial geometry, the original vehicle bounding-box annotations apply to all radiometric states.

\textbf{Shared Low-Rank Semantic Adaptation.} We feed $x_{s,\delta}$ into the DINOv3-LVD visual semantic branch and continue to feed the original image $x$ into the infrared spatial branch. To cover feature transformations at different depths while limiting the number of trainable parameters, we apply LoRA to the joint qkv mapping and attention output projection (proj) in the self-attention modules of transformer blocks 5, 12, 18, and 24. The computation is
\begin{equation}
\begin{aligned}h_{s,\delta}^{(l)}={}&W^{(l)}z_{s,\delta}^{(l)}+b^{(l)}\\ &+\frac{\alpha}{r}B^{(l)}A^{(l)}\operatorname{Dropout}(z_{s,\delta}^{(l)}).\end{aligned}
\label{eq:14}
\end{equation}
where $z_{s,\delta}^{(l)}$ and $h_{s,\delta}^{(l)}$ denote the input and output features, respectively, under radiometric state $(s,\delta)$ at LoRA insertion site $l$. The index $l$ identifies a specific transformer block and its $\mathrm{qkv}$ or $\mathrm{proj}$ mapping. The parameters $W^{(l)}$ and $b^{(l)}$ are frozen pretrained parameters, while $A^{(l)}$ and $B^{(l)}$ are the trainable low-rank matrices at that site. The symbol $r$ denotes the rank, $\alpha$ is the scaling coefficient, and $\operatorname{Dropout}(\cdot)$ denotes dropout in the low-rank update branch. Each insertion site shares the same low-rank parameters across radiometric states, allowing feature responses in different states to contribute to parameter updates through detection supervision.

Under this training scheme, the RI-LoRA optimization objective is the expected detection risk over the radiometric intervention distribution
\begin{equation}
\begin{aligned}\mathcal{L}_{\mathrm{RI}}(\Theta)={}&\mathbb{E}_{(x,y)\sim D_s}\mathbb{E}_{(s,\delta)\sim Q}\\ &\left[\mathcal{L}_{\mathrm{det}}\bigl(f_\Theta(T_{s,\delta}(x),x),y\bigr)\right].\end{aligned}
\label{eq:15}
\end{equation}
where $D_s$ is the joint distribution of source-domain training samples, $Q$ is the joint sampling distribution of radiometric intervention states and intensities, $f_\Theta$ is the two-branch detector, $\Theta$ denotes the parameters being optimized, and $L_{\mathrm{det}}$ is the detection loss detailed in Section~\ref{sec:fusion}.

In practice, we randomly sample one radiometric state per sample and use a single forward pass to obtain a Monte Carlo estimate of the expected detection risk
\begin{equation}
\widehat{L}_{\mathrm{RI}}=L_{\mathrm{det}}\bigl(f_\Theta(T_{s,\delta}(x)),y\bigr).
\label{eq:16}
\end{equation}

If the training samples and radiometric intervention states are sampled according to $D_s$ and $Q$, respectively, the sampling distributions are independent of the trainable parameters, and the usual differentiability and gradient--expectation interchange conditions hold, then
\begin{equation}
\begin{aligned}E[\widehat{L}_{\mathrm{RI}}]&=L_{\mathrm{RI}}(\Theta),\\ E[\nabla_\Theta\widehat{L}_{\mathrm{RI}}]&=\nabla_\Theta L_{\mathrm{RI}}(\Theta).\end{aligned}
\label{eq:17}
\end{equation}

These relations show that the sampled detection loss and its gradient are unbiased estimates of the expected detection risk and its gradient, respectively. Thus, a single forward pass suffices for stochastic optimization over the intervention distribution without simultaneously generating and processing multiple intervened views. Over training iterations, detection supervision under different radiometric states updates the shared low-rank parameters, extending semantic adaptation to varying radiometric conditions while retaining the original detection training procedure.

\subsection{SAOR: Semantic-Anchored Ordinal Relation Spatial Prior}\label{sec:saor}

\begin{figure}[!tb]
\centering
\includegraphics[width=\columnwidth,trim=19bp 19bp 19bp 12bp,clip]{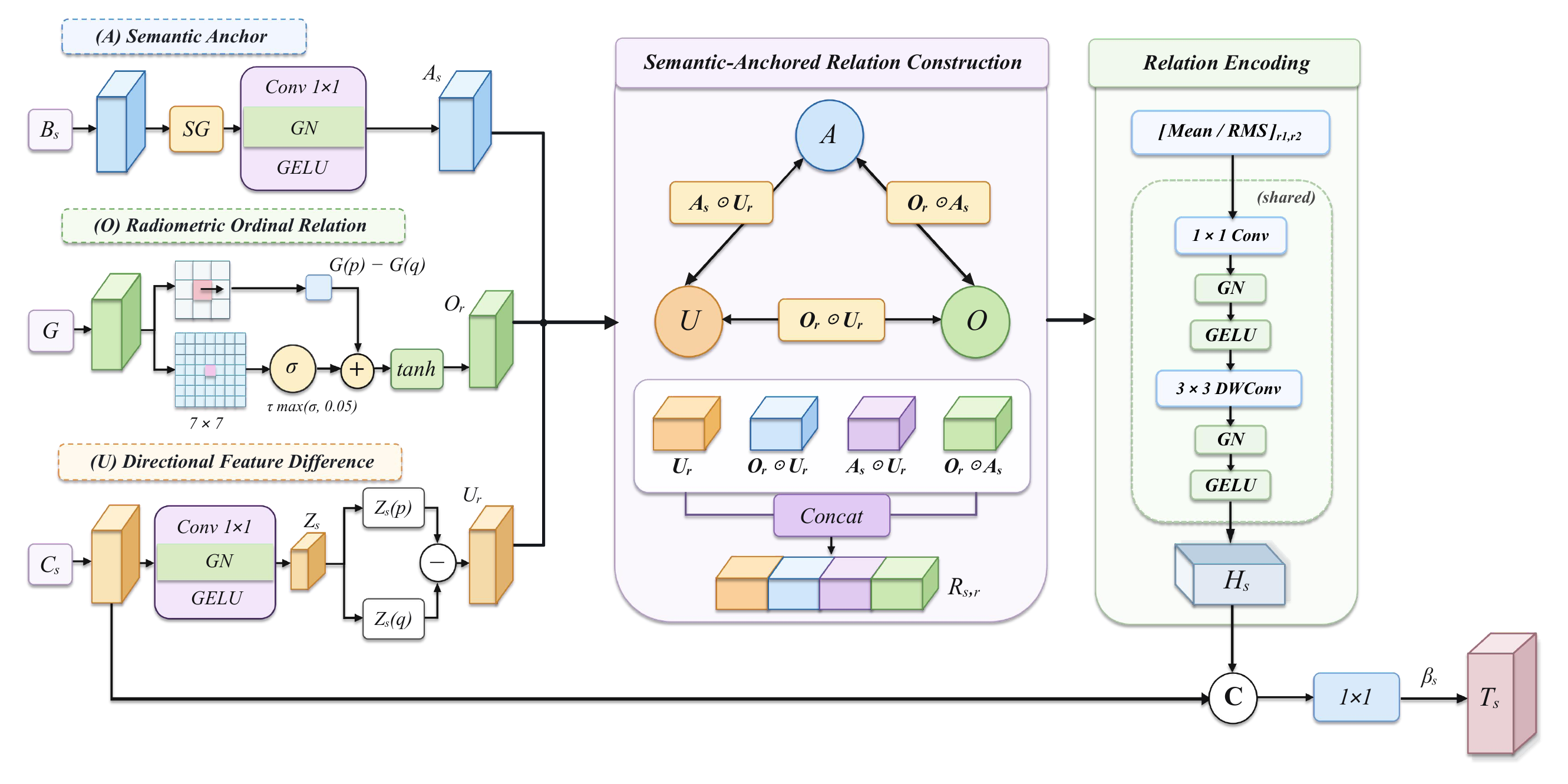}
\caption{Architecture of the semantic-anchored ordinal relation spatial prior (SAOR) module.}
\label{fig:2}
\end{figure}

In cross-scene infrared vehicle detection, local spatial features provide information about vehicle contours and boundaries. However, background structures such as road edges and building contours can also produce strong local responses, and radiometric differences between vehicles and backgrounds vary with imaging conditions. Local responses in convolutional features alone may therefore be insufficient for stable vehicle discrimination in complex backgrounds. SAOR combines signed radiometric ordinal relations, directional feature differences, and central semantic information to construct local spatial relations with a semantic reference. Fig.~\ref{fig:2} shows its overall architecture.

\textbf{Local Radiometric Ordinal Relations.} Infrared vehicles differ from their surroundings in both intensity magnitude and relative radiometric ordering across locations. To reduce the effect of variations in local contrast magnitude on the relation representation, SAOR normalizes center-to-neighbor radiometric differences using local statistics. We first average the input image across channels to obtain the intensity map $G$ and resize it to the corresponding feature scale. We compute statistics about center location $p$ within a $7\times7$ local window to obtain standard deviation $\sigma(p)$, then define the signed radiometric ordinal relation between center location $p$ and neighbor location $q$ as
\begin{equation}
O_s(p,q)=\tanh\left(\frac{G_s(p)-G_s(q)}{\tau\max(\sigma_s(p),\sigma_{\min})}\right),
\label{eq:18}
\end{equation}
where $\tau=0.35$ is a temperature coefficient that controls the response sensitivity to normalized radiometric differences, and the lower bound $\sigma_{\min}$ on the local standard deviation is 0.05. The sign encodes the radiometric ordering between the center and its neighbor, and the response magnitude represents the normalized radiometric difference.

Because $G_s$ provides only single-channel intensity information and cannot fully capture multichannel variations in convolutional features, we map the infrared convolutional features $C_s$ at the corresponding scale through a $1\times1$ convolution, group normalization $\mathrm{GN}$, and $\mathrm{GELU}$ activation to obtain 16-channel features $Z_s$. We then define the feature difference between the center and its neighbor as
\begin{equation}
U_{s(p,q)}=Z_{s(p)}-Z_{s(q)}.
\label{eq:19}
\end{equation}

Thus, $O_s$ describes the signed radiometric relation between the center and its neighbor, while $U_s$ preserves multichannel differences in feature space for that pair of locations. They encode relative radiometric variation and feature response differences, respectively, providing complementary information for subsequent relation encoding.

\textbf{Semantic-Anchored Joint Relation Construction.} Local radiometric variations and feature differences can also occur in nonvehicle regions. Pairwise differences alone may therefore be insufficient to characterize the underlying semantics. SAOR extracts information at the center from the high-level semantic features $B_s$ at the corresponding scale and projects it to the same channel dimension as $Z_s$ to form a semantic anchor
\begin{equation}
A_s=\operatorname{GELU}\bigl(\operatorname{GN}(\operatorname{Conv}_{1\times1}(\operatorname{sg}(B_s)))\bigr),
\label{eq:20}
\end{equation}
where $\operatorname{sg}(\cdot)$ denotes the stop-gradient operation, which blocks gradient propagation from the semantic anchor branch to upstream features while allowing the semantic projection layer to be trained. For a fixed center location $p$, semantic feature $A_s(p)$ is shared across its associated neighborhood relations and serves as their common semantic reference. SAOR then jointly models local feature differences, radiometric ordinal relations, and central semantics to construct the relation representation
\begin{equation}
\begin{aligned}R_s(p,q)=\operatorname{Concat}\bigl[&U_s(p,q),\\ &O_s(p,q)\odot U_s(p,q),\\ &A_s(p)\odot U_s(p,q),\\ &O_s(p,q)\odot A_s(p)\bigr].\end{aligned}
\label{eq:21}
\end{equation}
where $\odot$ denotes element-wise multiplication and the scalar radiometric ordinal relation is broadcast across channels. The four components preserve the original feature difference, modulate it with radiometric variation, associate the local difference with the central semantics, and associate the radiometric ordinal relation with the central semantics, respectively. Together they form a joint representation with a semantic reference.

\textbf{Multi-Range Relation Encoding and Spatial Prior Fusion.} Because local vehicle structures vary differently over different spatial ranges, SAOR constructs multi-range neighborhood relations at two high-resolution feature levels ($\mathit{stride}4$ and $\mathit{stride}8$). Each scale uses eight fixed neighborhood directions. The radius sets for $\mathit{stride}4$ and $\mathit{stride}8$ are \{1, 3\} and \{1, 2\}, respectively. For each radius, SAOR computes the mean and root-mean-square statistics of the joint relations over the eight directions, uses a shared relation encoder to extract local relation representations, and fuses the encodings from the two radii to obtain the spatial prior at the corresponding scale.

The spatial priors are injected into feature levels $\mathit{stride}4$ and $\mathit{stride}8$, while the other scales retain their original feature representations. The fused multiscale features are fed into the FPN and Faster R-CNN head for vehicle detection.

\subsection{Feature Fusion and Detection Objective}\label{sec:fusion}

Semantic adaptation and spatial relation modeling yield four levels of base semantic features $B=\{B_1,B_2,B_3,B_4\}$ and two levels of spatial priors $T_1,T_2$. As shown in Fig.~\ref{fig:1}(c), the spatial priors produced by SAOR are added element-wise to the first two base semantic feature levels, respectively, while the deeper levels retain their original semantic representations. This yields four levels of fused features
\begin{equation}
F_i=\begin{cases}B_i+T_i^{\mathrm{SAOR}},&i\in\{1,2\},\\B_i,&i\in\{3,4\}.\end{cases}
\label{eq:22}
\end{equation}

This fusion preserves the multiscale feature structure while incorporating semantic-anchored local relations into the high-resolution features to improve the spatial representations needed for vehicle recognition and localization.

As shown in Fig.~\ref{fig:1}(d), the four fused levels are fed into an FPN to generate multiscale feature maps for detection. The FPN follows the conventional feature pyramid design~\cite{ref41}. The RPN then generates region proposals, RoIAlign extracts their region features, and the RoI head predicts vehicle classes and regresses bounding boxes. The entire network is optimized using the standard Faster R-CNN detection loss
\begin{equation}
\mathcal{L}_{\mathrm{det}}=\mathcal{L}_{\mathrm{RPN}}^{\mathrm{cls}}+\mathcal{L}_{\mathrm{RPN}}^{\mathrm{reg}}+\mathcal{L}_{\mathrm{RoI}}^{\mathrm{cls}}+\mathcal{L}_{\mathrm{RoI}}^{\mathrm{reg}},
\label{eq:23}
\end{equation}
where $\mathcal{L}_{\mathrm{RPN}}^{\mathrm{cls}}$ and $\mathcal{L}_{\mathrm{RPN}}^{\mathrm{reg}}$ are the region proposal classification and bounding-box regression losses, respectively, and $\mathcal{L}_{\mathrm{RoI}}^{\mathrm{cls}}$ and $\mathcal{L}_{\mathrm{RoI}}^{\mathrm{reg}}$ are the RoI head classification and regression losses, respectively. This detection objective jointly optimizes RI-LoRA, SAOR, and the other trainable modules without auxiliary losses. The original pretrained backbone weights remain frozen.

During training, randomly intervened source-domain images are fed into the visual semantic branch, while the original images are fed into the infrared spatial branch. The original vehicle bounding-box annotations provide supervision. During inference, the radiometric intervention is disabled and detection uses the original images directly, without target-domain adaptation.

\section{Experiments}

\subsection{Datasets}

We use five thermal infrared vehicle detection datasets to evaluate cross-dataset generalization across different UAV-borne thermal infrared scenes. These include four public datasets, DroneVehicle~\cite{ref1}, ESCVehicle~\cite{ref7}, M3OT~\cite{ref42}, and HIT-UAV~\cite{ref4}, and VIVID, which was collected by our team. The datasets differ in imaging conditions, backgrounds, observation altitudes, vehicle scales, and per-image object density, providing complementary tests of detection performance in unseen scenes.

The infrared subset of DroneVehicle~\cite{ref1} contains 28,439 images, primarily covering urban roads, residential areas, and parking lots under daytime, nighttime, and low-illumination conditions. It provides rotated bounding-box annotations for five vehicle categories. The variation in vehicles and their spatial distributions across traffic scenes supports evaluation under different illumination conditions and complex urban traffic backgrounds.

The infrared subset of ESCVehicle~\cite{ref7} contains 10,727 images of roads, parking lots, built-up areas, and tree-covered regions. It includes rain, fog, snow-covered terrain, lakeside areas, and hills, and provides rotated bounding-box annotations for seven vehicle categories. Its varied weather conditions and natural land cover, together with structures such as vegetation and buildings, provide challenging scenarios for evaluating vehicle recognition and localization under environmental and background variation.

M3OT~\cite{ref42} contains 10,790 UAV-borne thermal infrared images acquired by two UAVs at observation altitudes of approximately 100--120 m. It covers urban and suburban scenes in daytime, at dusk, and at night. Vehicles are predominantly small because of the relatively high observation altitude, and differences in platforms and observation conditions introduce variation in object scale and viewing angle. This dataset provides an additional test of detection under high-altitude observation, small object scales, and diverse imaging conditions.

HIT-UAV~\cite{ref4} contains 2,898 UAV-borne thermal infrared images, primarily from schools, parking lots, roads, and sports fields. The images were extracted from hundreds of UAV videos and are accompanied by observation metadata, including flight altitude, camera angle, acquisition date, and sunlight intensity. The dataset provides both rotated and horizontal bounding boxes. We retain only the Car and OtherVehicle categories and merge them into the vehicle class. Variations in vehicle appearance due to flight conditions and viewing angles, and in the number of vehicles per image, complement the evaluation under different observation conditions and object distributions.

VIVID is a UAV-borne thermal infrared vehicle dataset collected by our team. We used two multirotor UAV platforms, the DJI Mavic 3T and DJI Matrice 4T, whose onboard thermal cameras acquired 640 \ensuremath{\times} 512 infrared images in the 8--14 \ensuremath{\mu}m band. The UAVs recorded low-altitude video of real traffic scenes. Frames were extracted and manually annotated with vehicle labels. The dataset includes vehicles moving in regular traffic and vehicles parked in the scene. It covers rural roads, buildings, vegetation, and other backgrounds, complementing the urban roads and parking lots commonly found in existing datasets. We constructed a dedicated evaluation set of 2,000 infrared images to assess cross-dataset generalization to small vehicles in complex land-cover backgrounds.

\begin{figure*}[!t]
\centering
\includegraphics[width=\textwidth,height=0.39\textheight,keepaspectratio]{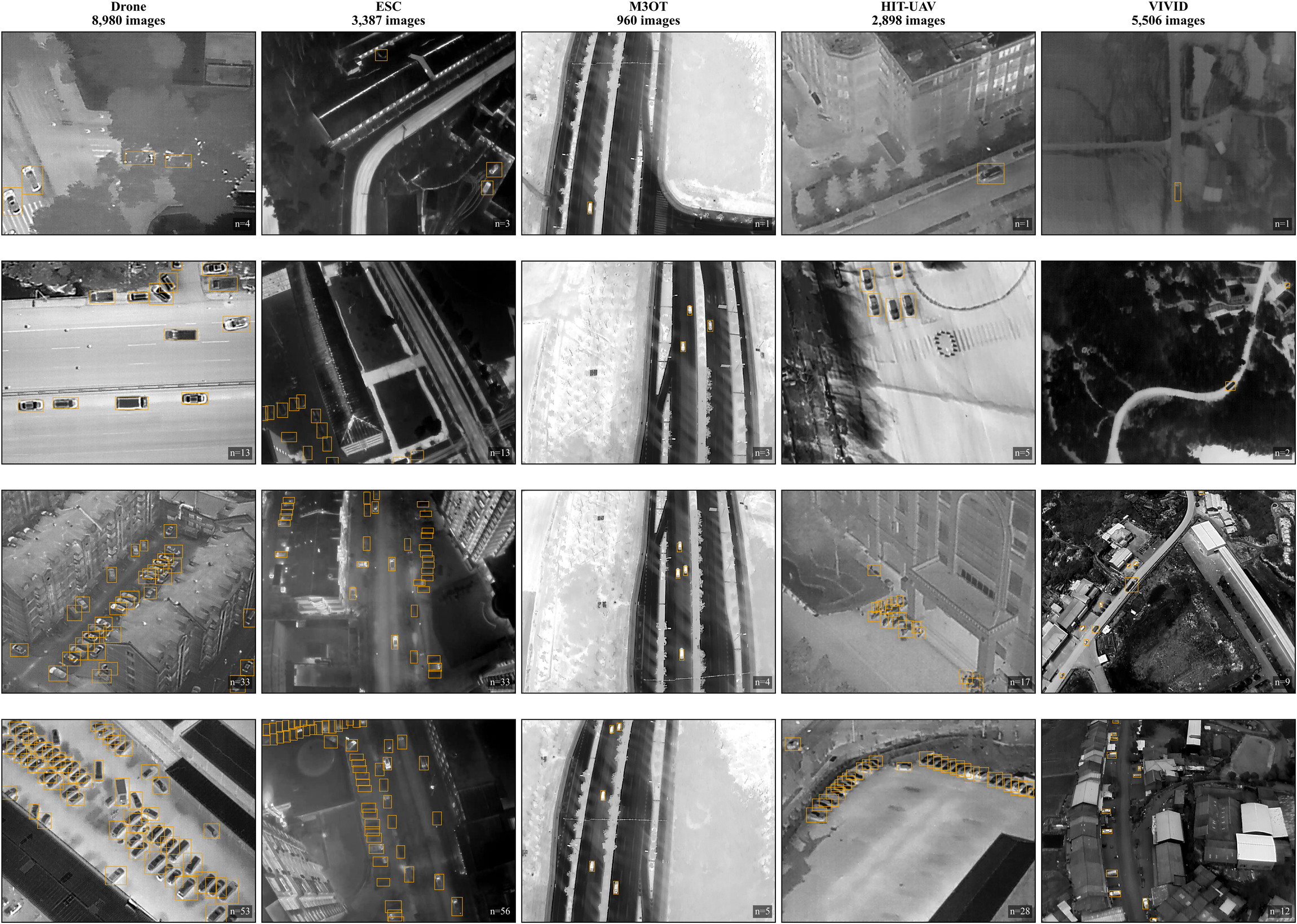}
\caption{Representative infrared images and vehicle bounding-box annotations from five datasets: DroneVehicle, ESCVehicle, M3OT, HIT-UAV, and VIVID.}
\label{fig:3}
\end{figure*}

Fig.~\ref{fig:3} shows representative infrared images and vehicle bounding-box annotations from the five datasets. Each column contains four samples from one dataset, ordered from top to bottom by increasing vehicle count. Comparisons across columns show differences in appearance, vehicle scale, and background composition, while comparisons within columns illustrate the range from sparse to dense vehicle distributions.

Fig.~\ref{fig:4} quantifies differences among the five datasets in image-level intensity statistics, spatial-frequency energy distribution, relative vehicle scale, and vehicles per image. Fig.~\ref{fig:4}(a) uses mean normalized intensity and intensity standard deviation to describe overall radiometric levels and intensity dispersion, respectively, revealing differences in overall intensity and within-image variation. Fig.~\ref{fig:4}(b) shows cumulative spectral energy as a function of normalized radial frequency, characterizing spatial-frequency composition and spectral energy distribution. Fig.~\ref{fig:4}(c) shows the distribution of the ratio of vehicle bounding-box area to image area. VIVID has a median relative vehicle area of 0.067\%, lower than the other four datasets, indicating a smaller typical relative vehicle scale. Fig.~\ref{fig:4}(d) compares object density through the distribution of vehicle counts per image. Together with Fig.~\ref{fig:3}, these comparisons show that VIVID includes smaller vehicles and diverse backgrounds such as rural roads, buildings, and vegetation. It thus extends cross-dataset generalization tests to small vehicles across different land-cover backgrounds, broadening coverage of object-scale variation and background complexity.

\subsection{Experimental Setup}

\subsubsection{Implementation and Model Configuration}

Existing research on domain-generalized object detection primarily addresses instance-level bounding boxes in visible-light images, whereas infrared cross-dataset research more often focuses on detecting small, dim targets and localizing their centers. Given these differences in task formulation and evaluation, we conduct separate comparisons for bounding-box and center-point detection. We compare representative general-purpose domain-generalized object detectors and cross-domain infrared small and dim target detectors, respectively, to assess vehicle instance detection and center localization.

We use two single-source protocols to assess cross-dataset generalization from different source domains. Both DroneVehicle and ESCVehicle use their official training and test splits. In Setting A, we train exclusively on the official DroneVehicle training split, evaluate in-domain (ID) on its official test split, and evaluate out-of-domain (OOD) on ESCVehicle, M3OT, HIT-UAV, and VIVID. In Setting B, we train exclusively on the official ESCVehicle training split, evaluate ID on its official test split, and evaluate OOD on DroneVehicle, M3OT, HIT-UAV, and VIVID. In both settings, all datasets other than the designated source are unseen target domains used only for final evaluation, with no role in training or parameter optimization.

We implement the method using PyTorch 2.1.2 and MMDetection 3.3.0 with CUDA 11.8. All experiments run on four NVIDIA RTX 4090 GPUs with 24 GB of memory each. We train for 16,000 iterations using SGD with an initial learning rate of 0.01, momentum of 0.9, weight decay of 1 \ensuremath{\times} \ensuremath{10^{-4}}, and total batch size of 8. The learning rate is warmed up linearly over the first 500 iterations and reduced to 0.1 times its initial value at iteration 13,000. We apply robust intensity normalization. Training augmentations include random resizing, cropping, and flipping. At test time, images are resized while preserving their original aspect ratio.

The visual backbone is DINOv3 ViT-L/16 initialized with LVD-1689M pretrained weights, which remain frozen. We also construct a DINOv2 ViT-L/14 variant to compare the effect of the pretrained backbone on detection. RI-LoRA and SAOR follow the architectures and parameter settings described in Sections~\ref{sec:ri-lora} and \ref{sec:saor}, respectively.

For comparability, we retrain all comparison methods on the datasets used in this study. Within each source-domain setting, all methods use the same source training data and the same ID and OOD test sets. Training and testing use the same hardware. Each method retains its original architecture and backbone configuration and is retrained with the training strategy and hyperparameters reported in its original publication. We use identical test data, class definitions, and metric computation protocols for all methods. We also use matched-backbone comparisons and controlled ablations to assess the contributions of the proposed modules.

\begin{figure}[!b]
\centering
\includegraphics[width=\columnwidth,height=0.48\textheight,keepaspectratio]{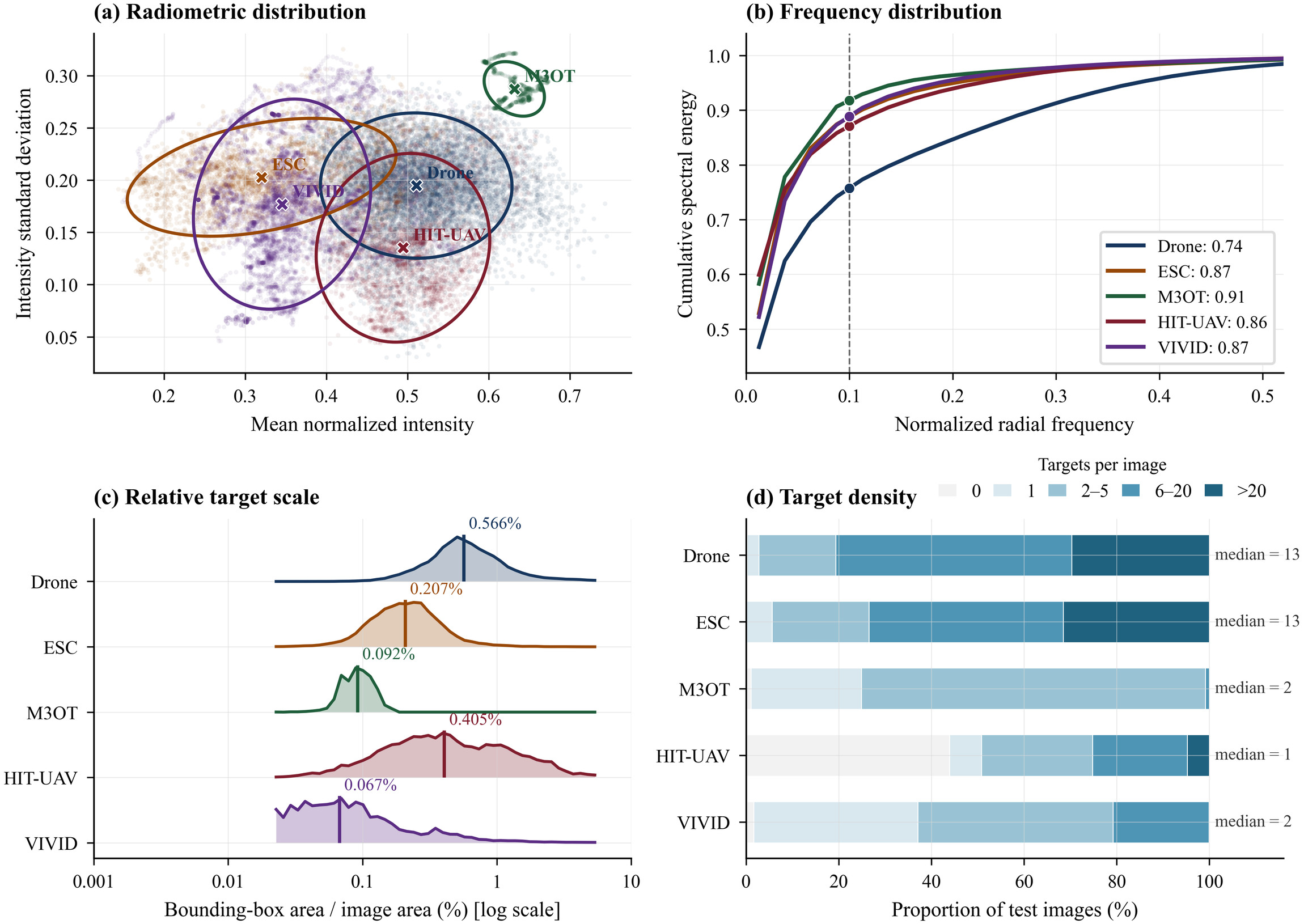}
\caption{Distribution characteristics of five aerial thermal infrared vehicle datasets: (a) image-level intensity statistics, (b) cumulative radial spectral energy, (c) relative vehicle scale, and (d) per-image vehicle count distribution.}
\label{fig:4}
\end{figure}

\subsubsection{Evaluation Metrics}

We quantify detection performance across datasets using metrics appropriate to bounding-box detection and center-point detection, respectively.

For bounding-box detection, we report average precision (AP50), precision, recall, and F1-score~\cite{ref43,ref44}. AP50 is the average precision at an intersection-over-union (IoU) threshold of 0.5. Precision and recall measure prediction accuracy and target detection, respectively. The F1-score is their harmonic mean.

For center-point detection, predicted boxes are converted to geometric centers, while response maps or segmentation results are converted to candidate centers with confidence scores. We use an adaptive distance-matching criterion based on the ground-truth box dimensions~\cite{ref45}
\begin{equation}
\lVert p_i-g_j\rVert_2\leq0.25\sqrt{w_jh_j},
\label{eq:24}
\end{equation}
where $p_i$ and $g_j$ denote the predicted and ground-truth centers, respectively, and $w_j$ and $h_j$ are the width and height of the ground-truth box. Predictions are ranked by confidence and matched one to one. We report center-point average precision (Point-AP), probability of detection (PD), and center localization error (CLE). Point-AP is computed from the precision--recall curve for center-point matching. PD is the fraction of ground-truth targets successfully detected, and CLE is the mean center-to-center distance, in pixels, for successful matches.

\subsection{Comparison with Representative Methods}

Under the experimental settings and evaluation protocols above, we compare methods on both bounding-box detection and center-point detection.

\subsubsection{Bounding-Box Detection}

We use Faster R-CNN~\cite{ref46} as the detection baseline and compare representative object detection and domain generalization methods, including OA-DG~\cite{ref24}, DivAlign~\cite{ref11}, SRCD~\cite{ref47}, PhysAug~\cite{ref28}, GDD~\cite{ref31}, and Bridge~\cite{ref32}. Tables~\ref{tab:1} and \ref{tab:2} report the results.

\begin{table*}[!tb]
\centering
\caption{In-Domain and Cross-Domain Bounding-Box Detection Results With DroneVehicle as the Source Domain}
\label{tab:1}
\footnotesize
\setlength{\tabcolsep}{1.7pt}
\renewcommand{\arraystretch}{1.10}

\begin{tabular*}{\textwidth}{@{\extracolsep{\fill}}ll*{12}{c}@{}}
\toprule
\textbf{Method} & \textbf{Backbone} & \multicolumn{2}{c}{\textbf{Drone (ID)}} & \multicolumn{2}{c}{\textbf{ESC (OOD)}} & \multicolumn{2}{c}{\textbf{M3OT (OOD)}} & \multicolumn{2}{c}{\textbf{HIT-UAV (OOD)}} & \multicolumn{2}{c}{\textbf{VIVID}\textbf{(OOD)}} & \multicolumn{2}{c}{\textbf{Avg. OOD}} \\
 &  & \textbf{AP50 $\uparrow$} & \textbf{F1 $\uparrow$} & \textbf{AP50 $\uparrow$} & \textbf{F1 $\uparrow$} & \textbf{AP50 $\uparrow$} & \textbf{F1 $\uparrow$} & \textbf{AP50 $\uparrow$} & \textbf{F1 $\uparrow$} & \textbf{AP50 $\uparrow$} & \textbf{F1 $\uparrow$} & \textbf{AP50 $\uparrow$} & \textbf{F1 $\uparrow$} \\
\midrule
Faster R-CNN & ResNet-50 & 96.00 & 93.23 & 36.66 & 43.36 & 38.03 & 26.02 & 57.68 & 60.61 & 13.38 & 22.60 & 36.44 & 38.15 \\
OA-DG \cite{ref24} & ResNet-101 & 96.80 & 94.18 & 35.30 & 44.47 & 66.50 & 54.60 & 54.10 & 60.88 & 13.20 & 20.88 & 42.28 & 45.21 \\
DivAlign \cite{ref11} & ResNet-101 & 96.40 & 91.27 & 40.50 & 55.36 & 49.20 & 56.68 & \textbf{59.80} & 63.13 & 13.20 & 23.18 & 40.68 & 49.59 \\
SRCD \cite{ref47} & ResNet-101 & 96.90 & 91.73 & 44.20 & 57.09 & 49.90 & 45.16 & 55.00 & 61.41 & 17.80 & 27.41 & 41.73 & 47.77 \\
PhysAug \cite{ref28} & ResNet-101 & 96.80 & 92.04 & 32.70 & 44.28 & 23.50 & 19.77 & 55.20 & 63.77 & 13.90 & 23.26 & 31.33 & 37.77 \\
GDD \cite{ref31} & SD-1.5 & 96.90 & 94.02 & 36.70 & 43.09 & 51.20 & 30.11 & 54.10 & 62.67 & 16.20 & 25.21 & 39.55 & 40.27 \\
Bridge \cite{ref32} & DINOv2-L/14 & 97.00 & 94.59 & 48.20 & 53.70 & \underline{95.70} & \underline{94.54} & 56.40 & 66.37 & 23.10 & 36.16 & 55.85 & 62.69 \\
\midrule
\textbf{Ours} & DINOv2-L/14 & \underline{97.10} & \underline{94.94} & \underline{53.80} & \underline{59.80} & 94.70 & 90.66 & \underline{57.90} & \underline{67.13} & \underline{24.90} & \underline{37.44} & \underline{57.83} & \underline{63.76} \\
 & DINOv3-L/16 & \textbf{97.20} & \textbf{95.01} & \textbf{57.80} & \textbf{62.64} & \textbf{96.20} & \textbf{95.09} & 56.60 & \textbf{67.49} & \textbf{25.40} & \textbf{39.63} & \textbf{59.00} & \textbf{66.21} \\
\bottomrule
\end{tabular*}
\par\medskip
\caption{In-Domain and Cross-Domain Bounding-Box Detection Results With ESCVehicle as the Source Domain}
\label{tab:2}
\footnotesize
\setlength{\tabcolsep}{1.7pt}
\renewcommand{\arraystretch}{1.10}

\begin{tabular*}{\textwidth}{@{\extracolsep{\fill}}ll*{12}{c}@{}}
\toprule
\textbf{Method} & \textbf{Backbone} & \multicolumn{2}{c}{\textbf{ESC (ID)}} & \multicolumn{2}{c}{\textbf{Drone (OOD)}} & \multicolumn{2}{c}{\textbf{M3OT (OOD)}} & \multicolumn{2}{c}{\textbf{HIT-UAV (OOD)}} & \multicolumn{2}{c}{\textbf{VIVID}\textbf{(OOD)}} & \multicolumn{2}{c}{\textbf{Avg. OOD}} \\
 &  & \textbf{AP50 $\uparrow$} & \textbf{F1 $\uparrow$} & \textbf{AP50 $\uparrow$} & \textbf{F1 $\uparrow$} & \textbf{AP50 $\uparrow$} & \textbf{F1 $\uparrow$} & \textbf{AP50 $\uparrow$} & \textbf{F1 $\uparrow$} & \textbf{AP50 $\uparrow$} & \textbf{F1 $\uparrow$} & \textbf{AP50 $\uparrow$} & \textbf{F1 $\uparrow$} \\
\midrule
Faster R-CNN & ResNet-50 & 84.66 & 81.04 & 78.18 & 75.05 & 96.44 & 75.21 & 51.95 & 53.27 & 31.39 & 39.63 & 64.49 & 60.79 \\
OA-DG \cite{ref24} & ResNet-101 & 77.00 & 80.02 & 85.50 & 85.06 & \textbf{98.40} & 91.34 & 56.90 & 60.72 & 40.90 & 52.02 & 70.43 & 72.29 \\
DivAlign \cite{ref11} & ResNet-101 & 64.80 & 63.38 & 76.50 & 65.95 & 87.90 & 68.60 & 51.60 & 46.94 & 25.60 & 37.09 & 60.40 & 54.65 \\
SRCD \cite{ref47} & ResNet-101 & 80.50 & 76.11 & 85.00 & 78.44 & 97.60 & 84.17 & 52.60 & 51.37 & 35.90 & 37.88 & 67.78 & 62.97 \\
PhysAug \cite{ref28} & ResNet-101 & 82.20 & 80.61 & 84.90 & 80.14 & 95.50 & 85.74 & 53.50 & 56.11 & 31.00 & 41.63 & 66.23 & 65.91 \\
GDD \cite{ref31} & SD-1.5 & 81.10 & 84.21 & 82.80 & 82.37 & 98.10 & 91.93 & 57.70 & 61.93 & 40.70 & 53.13 & 69.83 & 72.34 \\
Bridge \cite{ref32} & DINOv2-L/14 & 83.30 & 83.93 & 87.80 & 86.99 & 97.50 & \textbf{97.06} & \underline{57.90} & \underline{64.70} & 36.10 & 57.11 & 69.83 & 76.47 \\
\midrule
\textbf{Ours} & DINOv2-L/14 & \underline{85.50} & \underline{86.37} & \underline{88.50} & \underline{87.83} & \underline{98.30} & 96.40 & \textbf{58.90} & 64.41 & \textbf{45.20} & \underline{62.31} & \textbf{72.73} & \underline{77.74} \\
 & DINOv3-L/16 & \textbf{86.70} & \textbf{86.86} & \textbf{90.50} & \textbf{89.29} & \textbf{98.40} & \underline{96.61} & 57.60 & \textbf{64.73} & \underline{42.90} & \textbf{65.14} & \underline{72.35} & \textbf{78.94} \\
\bottomrule
\end{tabular*}
\end{table*}

Tables~\ref{tab:1} and \ref{tab:2} show that our method achieves competitive cross-dataset detection under both source-domain settings. With the same DINOv2-L/14 backbone as Bridge, our method improves mean OOD AP50/F1 by 1.98/1.07 and 2.90/1.27 percentage points with DroneVehicle and ESCVehicle as the source domains, respectively. This indicates that the framework improves cross-domain detection with the pretrained backbone held fixed. In particular, AP50/F1 improves by 5.60/6.10 and 9.10/5.20 percentage points on DroneVehicle\ensuremath{\to}ESCVehicle and ESCVehicle\ensuremath{\to}VIVID, respectively. The former gain reflects adaptation to different environmental imaging conditions, while the latter indicates improved detection of small vehicles in complex backgrounds.

With the DINOv3-L/16 backbone, our method achieves mean OOD AP50/F1 scores of 59.00\%/66.21\% and 72.35\%/78.94\% under the two source-domain settings, respectively. Together with the DINOv2 results, this shows that the framework performs well with both pretrained backbones and suggests that its gains also reflect the feature adaptation and spatial relation modeling designs. In particular, on DroneVehicle\ensuremath{\to}ESCVehicle, the DINOv3 variant reaches 57.80\% AP50 and 62.64\% F1, exceeding Bridge by 9.60 and 8.94 percentage points, respectively. This further demonstrates the framework's effectiveness in scenes with pronounced cross-domain performance degradation.

The bidirectional transfer results also show that source-domain selection affects cross-domain performance. For example, with the DINOv2 variant, training on ESCVehicle yields 45.20\% AP50 and 62.31\% F1 on VIVID, exceeding the results obtained with DroneVehicle as the source by 20.30 and 24.87 percentage points, respectively. In light of the radiometric statistics, spectral distributions, and object scales shown in Fig.~\ref{fig:4}, this transfer difference may be related to the environmental and imaging conditions covered by the source domain. ESCVehicle's diverse weather and natural land-cover backgrounds provide richer training conditions for learning discriminative vehicle features across radiometric states and background structures, offering a possible explanation for its transfer performance on complex scenes such as VIVID.

\subsubsection{Center-Point Detection}

We compare four representative cross-domain infrared target detectors, S2CPNet~\cite{ref48}, HSDW~\cite{ref49}, MMIDNet~\cite{ref50}, and SGIRNet~\cite{ref51}. Tables~\ref{tab:3} and \ref{tab:4} report the results.

\begin{table*}[!tb]
\centering
\caption{Cross-Domain Center-Point Detection Results With DroneVehicle as the Source Domain}
\label{tab:3}
\footnotesize
\setlength{\tabcolsep}{1.25pt}
\renewcommand{\arraystretch}{1.10}

\begin{tabular*}{\textwidth}{@{\extracolsep{\fill}}lc*{15}{c}@{}}
\toprule
\multicolumn{2}{c}{\textbf{Method}} & \multicolumn{3}{c}{\textbf{Drone (ID)}} & \multicolumn{3}{c}{\textbf{ESC (OOD)}} & \multicolumn{3}{c}{\textbf{M3OT (OOD)}} & \multicolumn{3}{c}{\textbf{HIT-UAV (OOD)}} & \multicolumn{3}{c}{\textbf{VIVID (OOD)}} \\
 &  & \textbf{Point-AP}~$\uparrow$ & \textbf{PD}~$\uparrow$ & \textbf{CLE}~$\downarrow$ & \textbf{Point-AP}~$\uparrow$ & \textbf{PD}~$\uparrow$ & \textbf{CLE}~$\downarrow$ & \textbf{Point-AP}~$\uparrow$ & \textbf{PD}~$\uparrow$ & \textbf{CLE}~$\downarrow$ & \textbf{Point-AP}~$\uparrow$ & \textbf{PD}~$\uparrow$ & \textbf{CLE}~$\downarrow$ & \textbf{Point-AP}~$\uparrow$ & \textbf{PD}~$\uparrow$ & \textbf{CLE}~$\downarrow$ \\
\midrule
\multicolumn{2}{l}{SGIRNet \cite{ref51}} & 88.31 & 85.26 & 2.67 & 13.44 & 23.71 & 3.30 & 5.14 & 17.04 & 2.53 & 40.50 & 60.49 & 3.68 & 5.46 & 15.40 & 3.03 \\
\multicolumn{2}{l}{MMIDNet \cite{ref50}} & 84.83 & 79.80 & 2.91 & 7.72 & 16.55 & 3.42 & 1.12 & 3.08 & 3.01 & 43.03 & 50.79 & 3.55 & 2.23 & 11.96 & 4.41 \\
\multicolumn{2}{l}{HSDW \cite{ref49}} & 87.97 & 83.88 & 2.58 & 13.56 & 22.97 & 3.29 & 10.32 & 27.55 & 2.29 & 40.09 & 57.16 & 3.49 & 7.04 & 14.98 & 2.90 \\
\multicolumn{2}{l}{S2CPNet \cite{ref48}} & 96.81 & 98.01 & 2.64 & 34.30 & 42.28 & 2.85 & 3.50 & 14.00 & 2.26 & 52.23 & \underline{63.62} & 3.63 & 11.66 & 22.33 & 3.20 \\
\midrule
\textbf{Ours} & DINOv2 & \underline{97.64} & \underline{98.63} & \underline{2.24} & \underline{56.37} & \underline{62.31} & \underline{2.44} & \underline{92.32} & \underline{95.46} & \textbf{1.48} & \textbf{58.78} & \textbf{63.78} & \textbf{2.35} & \underline{28.86} & \underline{39.67} & \underline{1.96} \\
\addlinespace
\textbf{Ours} & DINOv3 & \textbf{97.78} & \textbf{98.74} & \textbf{2.16} & \textbf{59.89} & \textbf{64.80} & \textbf{2.35} & \textbf{93.26} & \textbf{96.11} & \underline{1.65} & \underline{57.50} & 60.95 & \underline{2.39} & \textbf{31.76} & \textbf{43.51} & \textbf{1.93} \\
\bottomrule
\end{tabular*}
\par\medskip
\caption{Cross-Domain Center-Point Detection Results With ESCVehicle as the Source Domain}
\label{tab:4}
\footnotesize
\setlength{\tabcolsep}{1.25pt}
\renewcommand{\arraystretch}{1.10}

\begin{tabular*}{\textwidth}{@{\extracolsep{\fill}}lc*{15}{c}@{}}
\toprule
\multicolumn{2}{c}{\textbf{Method}} & \multicolumn{3}{c}{\textbf{ESC (ID)}} & \multicolumn{3}{c}{\textbf{Drone (OOD)}} & \multicolumn{3}{c}{\textbf{M3OT (OOD)}} & \multicolumn{3}{c}{\textbf{HIT-UAV (OOD)}} & \multicolumn{3}{c}{\textbf{VIVID (OOD)}} \\
 &  & \textbf{Point-AP}~$\uparrow$ & \textbf{PD}~$\uparrow$ & \textbf{CLE}~$\downarrow$ & \textbf{Point-AP}~$\uparrow$ & \textbf{PD}~$\uparrow$ & \textbf{CLE}~$\downarrow$ & \textbf{Point-AP}~$\uparrow$ & \textbf{PD}~$\uparrow$ & \textbf{CLE}~$\downarrow$ & \textbf{Point-AP}~$\uparrow$ & \textbf{PD}~$\uparrow$ & \textbf{CLE}~$\downarrow$ & \textbf{Point-AP}~$\uparrow$ & \textbf{PD}~$\uparrow$ & \textbf{CLE}~$\downarrow$ \\
\midrule
\multicolumn{2}{l}{SGIRNet \cite{ref51}} & 41.95 & 49.49 & 2.60 & 47.19 & 63.38 & 5.94 & 9.02 & 42.39 & 2.03 & 27.14 & 60.48 & 6.47 & 10.08 & 34.25 & 3.33 \\
\multicolumn{2}{l}{MMIDNet \cite{ref50}} & 45.36 & 43.56 & 2.32 & 53.33 & 50.73 & 4.13 & 42.74 & 64.26 & 1.69 & 34.42 & 51.83 & 4.81 & 14.33 & 30.79 & 2.97 \\
\multicolumn{2}{l}{HSDW \cite{ref49}} & 44.35 & 54.21 & 2.60 & 60.03 & 68.28 & 4.91 & 46.95 & 75.17 & 1.91 & 30.62 & 62.16 & 5.70 & 17.75 & 41.98 & 3.11 \\
\multicolumn{2}{l}{S2CPNet \cite{ref48}} & \textbf{86.14} & 90.01 & \underline{2.15} & 58.57 & 70.75 & 4.97 & 83.42 & 89.66 & 1.64 & 29.49 & 58.00 & 5.85 & 30.08 & 48.11 & 2.51 \\
\midrule
\textbf{Ours} & DINOv2 & 85.10 & \underline{90.03} & 2.17 & \underline{89.95} & \underline{93.94} & \underline{3.45} & \textbf{95.94} & \textbf{98.26} & \textbf{1.39} & \textbf{60.95} & \textbf{71.33} & \underline{3.31} & \underline{57.24} & \textbf{71.10} & \textbf{1.82} \\
\addlinespace
\textbf{Ours} & DINOv3 & \underline{86.01} & \textbf{90.48} & \textbf{2.09} & \textbf{91.34} & \textbf{95.09} & \textbf{3.15} & \underline{95.33} & \underline{98.01} & \underline{1.42} & \underline{59.23} & \underline{68.57} & \textbf{3.03} & \textbf{60.99} & \underline{71.01} & \underline{1.85} \\
\bottomrule
\end{tabular*}
\end{table*}

As shown in Table~\ref{tab:3}, with DroneVehicle as the source, some comparison methods achieve high in-domain accuracy but degrade substantially on unseen target domains. For example, the in-domain Point-AP of S2CPNet is 96.81\%, but falls to 34.30\% on ESCVehicle and 3.50\% on M3OT, indicating sensitivity to domain shifts. Our DINOv3 variant achieves 59.89\% Point-AP on ESCVehicle, 25.59 percentage points above the strongest comparison method, while also achieving higher PD and lower CLE. These results indicate improved target detection and center localization.

With ESCVehicle as the source, Table~\ref{tab:4} shows different transfer behavior. For example, the Point-AP of S2CPNet on M3OT rises to 83.42\%, markedly different from its performance under the other source-domain setting. This further illustrates the effect of source-domain selection. Our DINOv3 variant achieves 91.34\% Point-AP and 95.09\% PD on DroneVehicle, while the DINOv2 variant shows some advantages on M3OT and HIT-UAV. Overall, our method maintains good detection and localization across source settings and multiple unseen target domains.

Taken together, the two sets of experiments show that some comparison methods perform well in-domain but vary substantially in cross-dataset vehicle detection. This behavior is related to their original task focus. Existing methods for cross-domain UAV-borne infrared small and dim target detection emphasize weak target response extraction and background suppression, whereas vehicle center-point detection also depends on local vehicle structure and instance discrimination. Their original representations may therefore have limited applicability across vehicle scenes. Our method combines pretrained semantic representations, radiometric variation modeling, and local spatial relation modeling, yielding good detection and localization across transfer settings and demonstrating its applicability to cross-scene infrared vehicle detection.

\subsubsection{Qualitative Comparison of Bounding-Box Detection}

\begin{figure*}[!t]
\centering
\includegraphics[width=\textwidth]{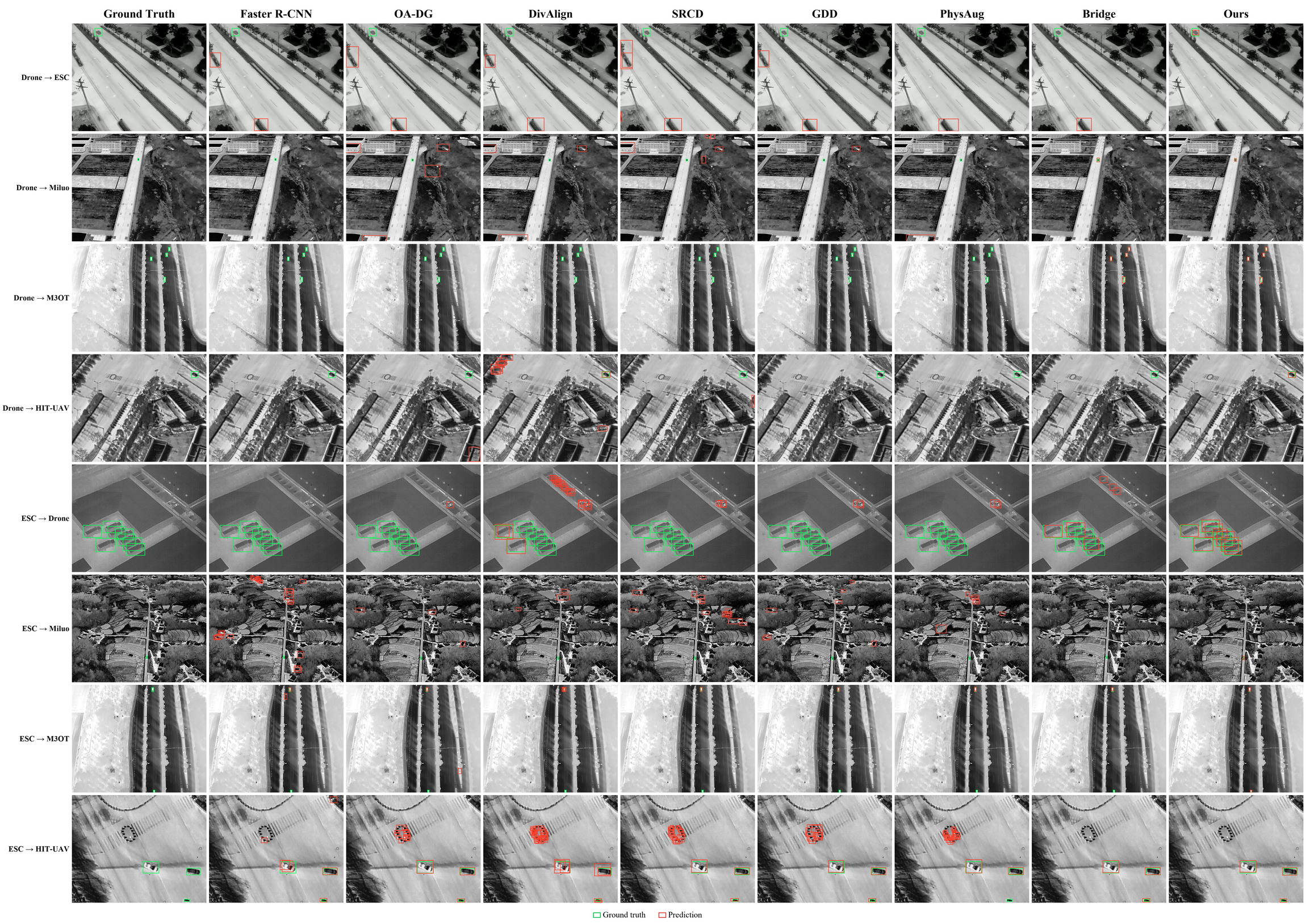}
\caption{Bounding-box predictions of different detection methods in infrared scenes.}
\label{fig:5}
\end{figure*}

Fig.~\ref{fig:5} shows bounding-box detections across datasets. Different background structures visibly interfere with vehicle detection. In ESCVehicle\ensuremath{\to}VIVID-20k, rural rooftops have pronounced edges and repetitive textures, and some methods misidentify the roofs and adjacent regions as vehicles. In ESCVehicle\ensuremath{\to}HIT-UAV, the ring-shaped ground structure in the upper left also produces false responses in several methods. In ESCVehicle\ensuremath{\to}DroneVehicle, a bright diagonal edge above the parking area produces additional predictions and makes it harder to distinguish vehicle instances accurately. In DroneVehicle\ensuremath{\to}ESCVehicle, some methods miss actual vehicles because of their small sizes and weak local responses. Our method localizes actual vehicles in these scenes while reducing spurious predictions from rooftops, ring-shaped structures, and road edges. In the densely populated ESCVehicle\ensuremath{\to}DroneVehicle example, it detects multiple adjacent vehicle instances more completely. These results show that the method improves small-vehicle detection and better distinguishes vehicles from similar-looking background structures, improving instance-level detection in complex infrared scenes.

\subsubsection{Qualitative Comparison of Center-Point Detection}

\begin{figure*}[!t]
\centering
\includegraphics[width=\textwidth,trim={4bp 6bp 10bp 9bp},clip]{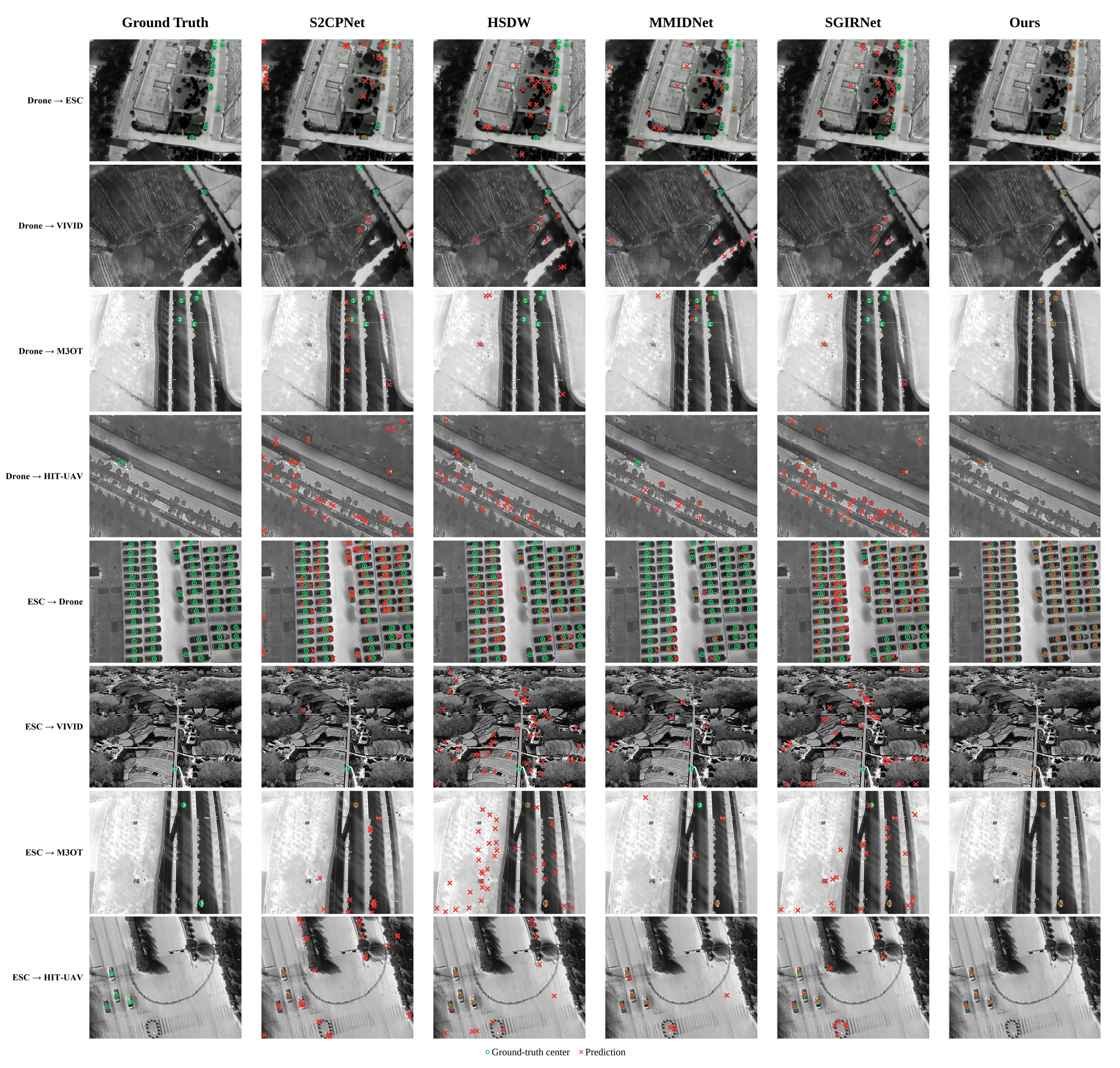}
\caption{Center-location predictions of different infrared object detection methods.}
\label{fig:6}
\end{figure*}

Fig.~\ref{fig:6} shows center-point detections across datasets. Green markers indicate ground-truth vehicle centers, and red markers indicate predicted centers. Some methods are susceptible to interference from nonvehicle structures and produce numerous scattered false predictions. In DroneVehicle\ensuremath{\to}HIT-UAV, for example, the continuous light--dark pattern formed by a road edge and adjacent vegetation produces dense false responses in several methods. In ESCVehicle\ensuremath{\to}VIVID-20k, rural rooftops and surrounding vegetation also produce many false predictions. In ESCVehicle\ensuremath{\to}M3OT, intensity transitions on both sides of the road likewise produce spurious responses. Our predictions are more concentrated at actual vehicle locations in these scenes, with fewer scattered false detections in the background. In the dense ESCVehicle\ensuremath{\to}DroneVehicle parking scene, our predicted centers correspond closely to the annotations and more fully preserve the center distribution of adjacent vehicles. These results show that the method reduces interference from roads, buildings, and vegetation while improving center matching in dense vehicle scenes, yielding more stable infrared vehicle center-point predictions across datasets.

\subsection{Ablation Studies}

To assess the effect of the proposed modules and their key design choices on cross-domain detection, we conduct ablations using DroneVehicle as the sole training source and ESCVehicle as the unseen test domain. All experiments use DINOv3 ViT-L/16 and identical training and detection configurations apart from the component under study. We compare RI-LoRA with vanilla LoRA and other radiometric intervention strategies to assess its effect on low-rank semantic adaptation. For SAOR, we use vanilla LoRA as the base adaptation configuration and progressively introduce spatial relation modeling and its key components to assess their individual contributions. Finally, we combine RI-LoRA and SAOR to examine their complementarity.

\subsubsection{Overall Component Ablation}

\begin{table}[!t]
\centering
\caption{Ablation Results for Different Model Components}
\label{tab:5}
\footnotesize
\setlength{\tabcolsep}{1.7pt}
\renewcommand{\arraystretch}{1.10}

\begin{tabular*}{\columnwidth}{@{\extracolsep{\fill}}*{7}{c}@{}}
\toprule
\textbf{CBB Base} & \textbf{RI-LoRA} & \textbf{SAOR} & \multicolumn{4}{c}{\textbf{Performance}} \\
 &  &  & \textbf{Prec.}~$\uparrow$ & \textbf{Recall}~$\uparrow$ & \textbf{F1}~$\uparrow$ & \textbf{AP50}~$\uparrow$ \\
\midrule
$\times$ & $\times$ & $\times$ & 96.25 & 38.37 & 54.87 & 51.8 \\
$\checkmark$ & $\times$ & $\times$ & \textbf{96.53} & 42.72 & 59.23 & 53.4 \\
$\checkmark$ & $\checkmark$ & $\times$ & 96.03 & 45.00 & 61.28 & 56.4 \\
$\checkmark$ & $\times$ & $\checkmark$ & 96.34 & 44.98 & 61.33 & 56.6 \\
$\checkmark$ & $\checkmark$ & $\checkmark$ & 96.20 & \textbf{46.44} & \textbf{62.64} & \textbf{57.8} \\
\bottomrule
\end{tabular*}
\end{table}

Table~\ref{tab:5} summarizes cross-domain detection for different module combinations. Adding CBB Base to the baseline improves AP50 and recall by 1.60 and 4.35 percentage points, respectively. Adding RI-LoRA or SAOR separately to this configuration yields AP50 values of 56.40\% and 56.60\%, respectively, and recall values of 45.00\% and 44.98\%, respectively, indicating that both modules improve vehicle detection. Combining the two yields 57.80\% AP50, 46.44\% recall, and a 62.64\% F1-score. AP50 improves by 1.40 and 1.20 percentage points over using RI-LoRA and SAOR individually, respectively. Precision remains near 96\% across configurations, indicating that the gains primarily reflect improved vehicle detection without an appreciable drop in prediction precision. The two modules are effective individually and complement each other when combined.

\subsubsection{RI-LoRA Ablation}

To assess the effect of the radiometric intervention strategy on cross-domain detection, we compare standard LoRA (w/o RI), Fourier amplitude mixing (FAM), and full RI-LoRA, holding the LoRA insertion sites, rank, and training configuration fixed. Table~\ref{tab:6} reports the results.

Table~\ref{tab:6} compares detection performance across the three configurations. Compared with standard LoRA, FAM improves AP50, recall, and F1-score by 1.10, 0.76, and 0.84 percentage points, respectively. Full RI-LoRA further increases these metrics to 56.40\%, 45.00\%, and 61.28\%, respectively, for additional gains of 0.70, 0.58, and 0.44 percentage points over FAM. Precision remains above 95\% in all three configurations, suggesting that the gains primarily reflect improved vehicle detection. These results indicate that general frequency-domain amplitude variation improves cross-domain feature adaptation, and targeted intervention in low-frequency radiometric statistics yields additional gains.

\begin{figure}[t]
\centering
\includegraphics[width=.42\columnwidth,trim=153bp 190bp 156bp 4bp,clip]{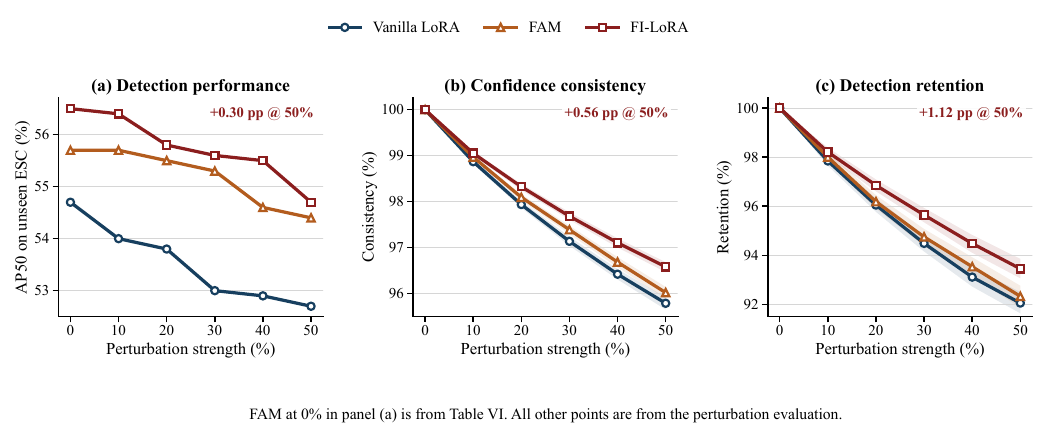}\par\smallskip
\makebox[\columnwidth][c]{%
\includegraphics[width=.333\columnwidth,trim=5bp 34bp 338bp 31bp,clip]{figures/fig07.pdf}%
\includegraphics[width=.333\columnwidth,trim=170bp 34bp 173bp 31bp,clip]{figures/fig07.pdf}%
\includegraphics[width=.333\columnwidth,trim=338bp 34bp 5bp 31bp,clip]{figures/fig07.pdf}%
}
\caption{Cross-domain detection performance and prediction stability under low-frequency amplitude perturbations. (a) AP50 across perturbation strengths; (b) prediction confidence consistency; (c) retention rate of correctly detected objects. Shaded regions denote 95\% confidence intervals estimated by image-level bootstrap. FAM at 0\% in panel (a) is from Table~\ref{tab:6}. All other points are from the perturbation evaluation.}
\label{fig:7}
\end{figure}

To assess how well each method maintains performance under changes in input radiometric statistics, we apply low-frequency amplitude perturbations of varying strengths to the unseen ESCVehicle test set and evaluate overall detection performance and prediction stability. Fig.~\ref{fig:7} presents the results. AP50 measures overall detection performance. Prediction confidence consistency and the retention rate of correctly detected targets before and after perturbation respectively measure how well prediction responses and originally correct detections are preserved.

\par\medskip
\noindent\begin{minipage}{\columnwidth}
\expandafter\def\csname @captype\endcsname{table}
\centering
\caption{Ablation Results for RI-LoRA Components on ESCVehicle}
\label{tab:6}
\footnotesize
\setlength{\tabcolsep}{1.7pt}
\renewcommand{\arraystretch}{1.10}

\begin{tabular*}{\linewidth}{@{\extracolsep{\fill}}l*{4}{c}@{}}
\toprule
\textbf{Method} & \textbf{Prec.}~$\uparrow$ & \textbf{Recall}~$\uparrow$ & \textbf{F1}~$\uparrow$ & \textbf{AP50}~$\uparrow$ \\
\midrule
w/o RI (Vanilla LoRA) & 95.87 & 43.66 & 60.00 & 54.60 \\
w/ FAM instead of RI & \textbf{96.56} & 44.42 & 60.84 & 55.70 \\
Full RI-LoRA & 96.03 & \textbf{45.00} & \textbf{61.28} & \textbf{56.40} \\
\bottomrule
\end{tabular*}
\end{minipage}\par\medskip

As shown in Fig.~\ref{fig:7}, AP50, confidence consistency, and target retention decrease for all three methods as low-frequency amplitude perturbation increases. RI-LoRA leads at every nonzero perturbation strength. At 50\% perturbation strength, it retains 54.70\% AP50, 0.30 percentage points above FAM. Its prediction confidence consistency is 96.60\%, 0.56 percentage points above FAM, and its correctly detected target retention is 93.50\%, 1.12 percentage points above FAM. In particular, the higher retention rate indicates that RI-LoRA preserves more originally correct vehicle detections when radiometric responses change, reflecting greater detection stability.

Together, Table~\ref{tab:6} and Fig.~\ref{fig:7} show that RI-LoRA improves standard cross-domain metrics and helps preserve prediction confidence and correct detections under changes in radiometric statistics. Compared with general Fourier amplitude mixing, targeted low-frequency radiometric intervention further improves adaptation to different radiometric states, supporting the effectiveness of RI-LoRA for cross-domain feature adaptation and prediction stability.

\subsubsection{SAOR Ablation}

To assess the effect of the semantic anchor and signed radiometric ordinal relations on cross-domain detection, we evaluate four configurations while holding the other training settings fixed. These comprise no SAOR (w/o SAOR), SAOR without the semantic anchor (w/o Semantic Anchor), SAOR without the ordinal relation (w/o Ordinal Relation), and full SAOR. Table~\ref{tab:7} reports the results.

Table~\ref{tab:7} shows that full SAOR achieves the highest recall, F1-score, and AP50 at 44.98\%, 61.33\%, and 56.60\%, respectively. It reaches 96.34\% precision, tied for the highest value with the configuration without the semantic anchor. Compared with the configuration without SAOR, the full module improves AP50, recall, and F1-score by 2.00, 1.32, and 1.33 percentage points, respectively. This indicates that local spatial relation modeling improves vehicle detection while maintaining high precision.

\par\medskip
\noindent\begin{minipage}{\columnwidth}
\expandafter\def\csname @captype\endcsname{table}
\centering
\caption{Ablation Results for SAOR Components on ESCVehicle}
\label{tab:7}
\footnotesize
\setlength{\tabcolsep}{1.7pt}
\renewcommand{\arraystretch}{1.10}

\begin{tabular*}{\linewidth}{@{\extracolsep{\fill}}l*{4}{c}@{}}
\toprule
\textbf{Method} & \textbf{Prec.}~$\uparrow$ & \textbf{Recall}~$\uparrow$ & \textbf{F1}~$\uparrow$ & \textbf{AP50}~$\uparrow$ \\
\midrule
w/o SAOR & 95.87 & 43.66 & 60.00 & 54.60 \\
w/o Semantic Anchor & \textbf{96.34} & 43.65 & 60.08 & 55.70 \\
w/o Ordinal Relation & 96.08 & 40.88 & 57.36 & 55.20 \\
Full SAOR & \textbf{96.34} & \textbf{44.98} & \textbf{61.33} & \textbf{56.60} \\
\bottomrule
\end{tabular*}
\end{minipage}\par\medskip

Removing the semantic anchor reduces AP50, recall, and F1-score by 0.90, 1.33, and 1.25 percentage points, respectively, while precision is unchanged. This suggests that the anchor provides a high-level semantic reference for local radiometric relations at the center. Linking local variations with vehicle semantics helps distinguish vehicles from similar background structures such as buildings and vegetation.

Removing the signed radiometric ordinal relation reduces AP50, recall, and F1-score by 1.40, 4.10, and 3.97 percentage points, respectively, with a particularly pronounced decrease in recall. This suggests that the signed radiometric ordinal relation is important for vehicle detection. By encoding relative center-to-neighbor radiometric ordering and the magnitude of normalized differences, it preserves both the direction and relative strength of local radiometric changes. This provides more discriminative spatial relation information for local vehicle structures and improves detection of adjacent vehicles and vehicles in complex backgrounds.

To examine how the components affect local spatial responses, Fig.~\ref{fig:8} visualizes spatial relation contributions for SAOR without the semantic anchor, SAOR without the ordinal relation, and full SAOR.

\par\medskip
\noindent\begin{minipage}{\columnwidth}
\expandafter\def\csname @captype\endcsname{figure}
\centering
\includegraphics[width=\linewidth]{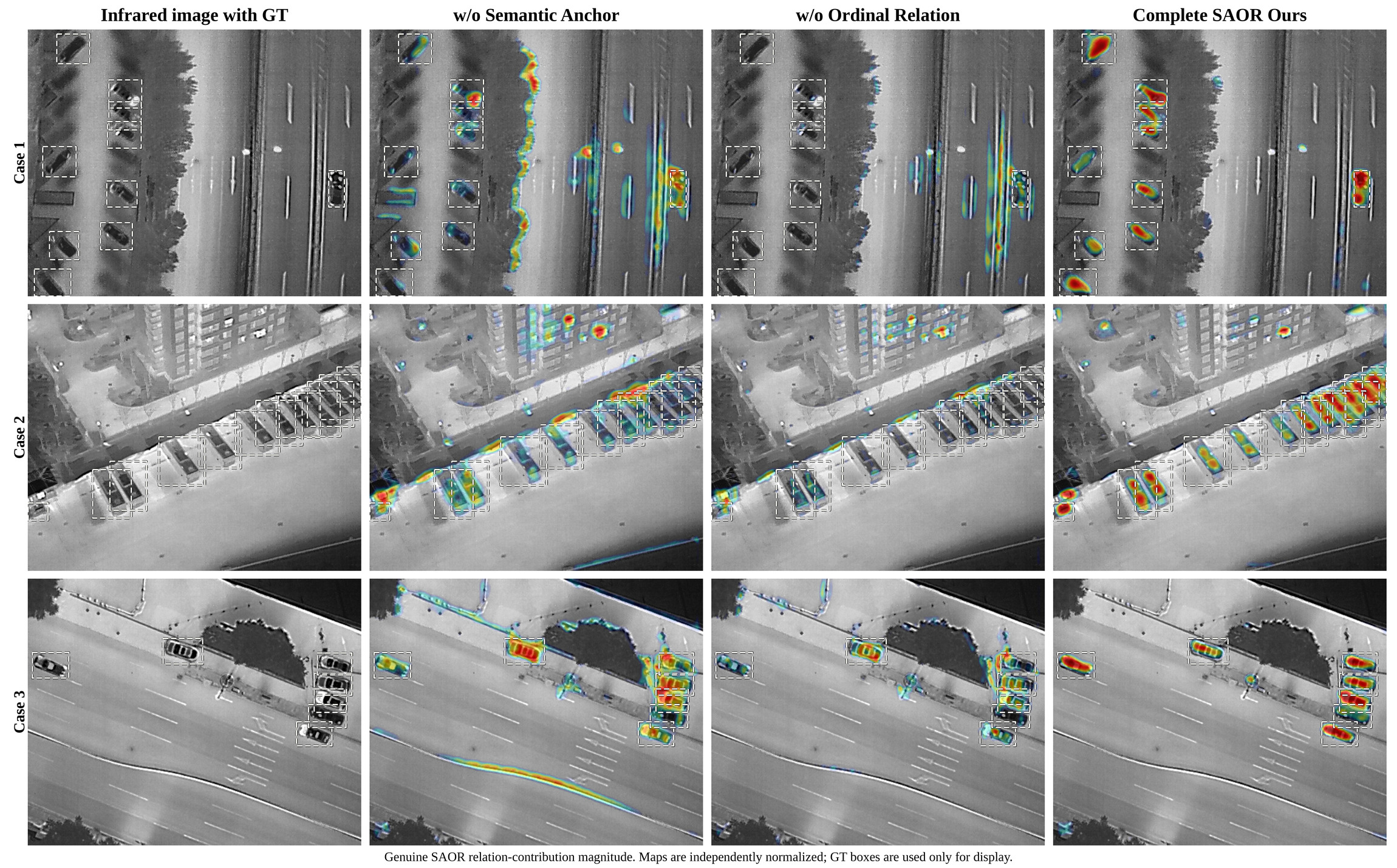}
\caption{Visualization of spatial relation contributions under different SAOR configurations. From left to right: infrared images with ground-truth bounding boxes, SAOR without the semantic anchor, SAOR without the ordinal relation, and the complete SAOR module.}
\label{fig:8}
\end{minipage}\par\medskip

In Fig.~\ref{fig:8}, removing the semantic anchor produces pronounced responses at the vegetation--road boundary in Case 1, at the windowpanes on the building fa\c{c}ade and the diagonal building--road boundary in Case 2, and at the road--vegetation boundary in Case 3. Removing the signed radiometric ordinal relation makes responses more dispersed and less complete within the parked vehicles on the left in Case 1 and the diagonally arranged vehicles in Case 2. Full SAOR concentrates responses in vehicle regions, weakens continuous responses along background boundaries, and more fully represents the local structures of adjacent vehicles. This suggests that semantic anchors help suppress interference from structures such as buildings and vegetation, while signed radiometric ordinal relations improve the representation of local vehicle structures. These observations are consistent with the quantitative results in Table~\ref{tab:7}.

\section{Conclusion}

This study addresses cross-dataset generalization of UAV-borne thermal infrared vehicle detection under single-source training. It focuses on shifts in feature adaptation caused by changes in radiometric statistics and on distinguishing local vehicle structures from similar structures in complex backgrounds. We propose IRadiDINO, a domain generalization framework based on a vision foundation model, to address these related issues. Building on pretrained visual representations, RI-LoRA exposes feature adaptation to a wider range of radiometric states and reduces dependence on source-specific radiometric appearances. SAOR incorporates local relative radiometric variations and pretrained semantics into spatial relation modeling to distinguish vehicle structures from complex backgrounds. The two components improve cross-scene detection features through adaptation to radiometric variations and representation of local spatial relations, respectively, providing complementary mechanisms for domain generalization.

We establish a bidirectional cross-dataset evaluation protocol spanning five datasets, comprising four public infrared vehicle datasets and VIVID, which we collected. We evaluate both bounding-box detection and center-point localization. IRadiDINO generalizes across source-domain settings and pretrained backbones and outperforms representative comparison methods under matched backbone conditions. Ablation, radiometric perturbation, and spatial response analyses further indicate that RI-LoRA improves feature adaptation and prediction stability across radiometric states, while SAOR improves the representation of local vehicle structures and their discrimination from complex backgrounds. Combining them further improves cross-domain detection.

Overall, cross-dataset generalization for UAV-borne thermal infrared vehicle detection requires accounting for the effect of scene-dependent radiometric statistics on feature representations and distinguishing local vehicle structures from complex backgrounds. Combining radiometric state intervention and semantic-anchored local spatial relation modeling provides an approach to both problems and offers a basis for adapting vision foundation models to cross-scene UAV-borne thermal infrared vehicle detection. Future work will extend the evaluation to additional imaging devices, radiometric conditions, and unseen scenes and investigate feature adaptation and spatial discrimination under broader imaging variations.


\begin{thebibliography}{99}

\bibitem{ref1} Y. Sun, B. Cao, P. Zhu, and Q. Hu, ``Drone-Based RGB-Infrared Cross-Modality Vehicle Detection via Uncertainty-Aware Learning,'' IEEE Trans. Circuits Syst. Video Technol., vol. 32, no. 10, pp. 6700--6713, Oct. 2022, doi: 10.1109/TCSVT.2022.3168279.

\bibitem{ref2} F. Outay, H. A. Mengash, and M. Adnan, ``Applications of Unmanned Aerial Vehicle (UAV) in Road Safety, Traffic and Highway Infrastructure Management: Recent Advances and Challenges,'' Transp. Res. Part A: Policy Pract., vol. 141, pp. 116--129, Nov. 2020, doi: 10.1016/j.tra.2020.09.018.

\bibitem{ref3} S. Srivastava, S. Narayan, and S. Mittal, ``A Survey of Deep Learning Techniques for Vehicle Detection from UAV Images,'' J. Syst. Archit., vol. 117, Art. no. 102152, Aug. 2021, doi: 10.1016/j.sysarc.2021.102152.

\bibitem{ref4} J. Suo, T. Wang, X. Zhang, H. Chen, W. Zhou, and W. Shi, ``HIT-UAV: A High-Altitude Infrared Thermal Dataset for Unmanned Aerial Vehicle-Based Object Detection,'' Sci. Data, vol. 10, no. 1, Art. no. 227, Apr. 2023, doi: 10.1038/s41597-023-02066-6.

\bibitem{ref5} N. Zhang, Y. Liu, H. Liu, T. Tian, J. Ma, and J. Tian, ``DTNet: A Specialized Dual-Tuning Network for Infrared Vehicle Detection in Aerial Images,'' IEEE Trans. Geosci. Remote Sens., vol. 62, Art. no. 5002815, 2024, doi: 10.1109/TGRS.2024.3386309.

\bibitem{ref6} S. Becker, A.-K. Grosselfinger, J. Bayer, D. M\"{u}nch, W. H\"{u}bner, and M. Arens, ``Fusion or Confusion? A Look at Dataset Pooling for Infrared Object Detection,'' in Proc. IEEE/CVF Conf. Comput. Vis. Pattern Recognit. Workshops (CVPRW), 2025, pp. 4471--4480, doi: 10.1109/CVPRW67362.2025.00428.

\bibitem{ref7} J. Song, N. Zhang, Z. Wang, and T. Tian, ``ESCVehicle: A Drone-Based Visible-Infrared Vehicle Benchmark With Extensive Scene Coverage,'' IEEE Trans. Geosci. Remote Sens., vol. 64, pp. 1--16, 2026, doi: 10.1109/TGRS.2026.3655959.

\bibitem{ref8} H.-K. Hsu et al., ``Progressive Domain Adaptation for Object Detection,'' in Proc. IEEE Winter Conf. Appl. Comput. Vis. (WACV), 2020, pp. 738--746.

\bibitem{ref9} C. Agnew, C. Eising, P. Denny, A. G. Scanlan, P. van de Ven, and E. M. Grua, ``Quantifying the Effects of Ground Truth Annotation Quality on Object Detection and Instance Segmentation Performance,'' IEEE Access, vol. 11, pp. 25174--25188, 2023, doi: 10.1109/ACCESS.2023.3256723.

\bibitem{ref10} V. Vidit, M. Engilberge, and M. Salzmann, ``CLIP the Gap: A Single Domain Generalization Approach for Object Detection,'' in Proc. IEEE/CVF Conf. Comput. Vis. Pattern Recognit. (CVPR), 2023, pp. 3219--3229, doi: 10.1109/CVPR52729.2023.00314.

\bibitem{ref11} M. S. Danish, M. H. Khan, M. A. Munir, M. S. Sarfraz, and M. Ali, ``Improving Single Domain-Generalized Object Detection: A Focus on Diversification and Alignment,'' in Proc. IEEE/CVF Conf. Comput. Vis. Pattern Recognit. (CVPR), 2024, pp. 17732--17742, doi: 10.1109/CVPR52733.2024.01679.

\bibitem{ref12} Y. Liu, S. Zhou, X. Liu, C. Hao, B. Fan, and J. Tian, ``Unbiased Faster R-CNN for Single-Source Domain Generalized Object Detection,'' in Proc. IEEE/CVF Conf. Comput. Vis. Pattern Recognit. (CVPR), 2024, pp. 28838--28847, doi: 10.1109/CVPR52733.2024.02724.

\bibitem{ref13} Y. Lu, Y. Li, X. Guo, S. Yuan, Y. Shi, and L. Lin, ``Rethinking Generalizable Infrared Small Target Detection: A Real-Scene Benchmark and Cross-View Representation Learning,'' IEEE Trans. Geosci. Remote Sens., vol. 63, pp. 1--10, 2025, doi: 10.1109/TGRS.2025.3588885.

\bibitem{ref14} R. Kou et al., ``Infrared Small Target Segmentation Networks: A Survey,'' Pattern Recognit., vol. 143, Art. no. 109788, 2023, doi: 10.1016/j.patcog.2023.109788.

\bibitem{ref15} M. Awais et al., ``Foundation Models Defining a New Era in Vision: A Survey and Outlook,'' IEEE Trans. Pattern Anal. Mach. Intell., vol. 47, no. 4, pp. 2245--2264, Apr. 2025, doi: 10.1109/TPAMI.2024.3506283.

\bibitem{ref16} M. Oquab et al., ``DINOv2: Learning Robust Visual Features without Supervision,'' Trans. Mach. Learn. Res., 2024. [Online]. Available: \url{https://arxiv.org/abs/2304.07193}

\bibitem{ref17} D.-G. Lee, J. Kim, Y. Cho, and A. Kim, ``Thermal Chameleon: Task-Adaptive Tone-Mapping for Radiometric Thermal-Infrared Images,'' IEEE Robot. Autom. Lett., vol. 9, no. 12, pp. 10827--10834, Dec. 2024, doi: 10.1109/LRA.2024.3479700.

\bibitem{ref18} J. Liu, Y. Xie, and J. Wang, ``UniScale-DETR: A Unified Transformer Framework for Small Object Detection in UAV Thermal Infrared Imagery,'' Infrared Phys. Technol., vol. 159, Art. no. 106856, 2026, doi: 10.1016/j.infrared.2026.106856.

\bibitem{ref19} C. Yu et al., ``Research on Vehicle Detection in Infrared Aerial Images in Complex Urban and Road Backgrounds,'' Electronics, vol. 13, no. 2, Art. no. 319, 2024, doi: 10.3390/electronics13020319.

\bibitem{ref20} Y. Zhang, Z. Dai, C. Pan, G. Zhang, and J. Xu, ``NOC-YOLO: An Exploration to Enhance Small-Target Vehicle Detection Accuracy in Aerial Infrared Images,'' Infrared Phys. Technol., vol. 149, Art. no. 105905, 2025, doi: 10.1016/j.infrared.2025.105905.

\bibitem{ref21} Q. Wang, F. Liu, Y. Cao, F. Ullah, and M. Zhou, ``LFIR-YOLO: Lightweight Model for Infrared Vehicle and Pedestrian Detection,'' Sensors, vol. 24, no. 20, Art. no. 6609, 2024, doi: 10.3390/s24206609.

\bibitem{ref22} N. Zhang et al., ``Omni-Scene Infrared Vehicle Detection: An Efficient Selective Aggregation Approach and a Unified Benchmark,'' ISPRS J. Photogramm. Remote Sens., vol. 223, pp. 244--260, 2025, doi: 10.1016/j.isprsjprs.2025.03.002.

\bibitem{ref23} B. Pei, L. Wu, X. Zheng, C. Zhou, and D. Wang, ``Fs2PA: A Full-Scale Feature Synergistic Perception Architecture for Vehicular Infrared Object Detection via Physical Priors and Semantic Constraints,'' Sensors, vol. 26, no. 7, Art. no. 2257, 2026, doi: 10.3390/s26072257.

\bibitem{ref24} W. Lee, D. Hong, H. Lim, and H. Myung, ``Object-Aware Domain Generalization for Object Detection,'' in Proc. AAAI Conf. Artif. Intell., vol. 38, no. 4, pp. 2947--2955, 2024, doi: 10.1609/aaai.v38i4.28076.

\bibitem{ref25} C. Lu, S. Su, and H. Li, ``Statistical-Driven Adaptive Data Augmentation for Single-Domain Generalized Object Detection,'' Comput. Vis. Image Underst., vol. 265, Art. no. 104687, 2026, doi: 10.1016/j.cviu.2026.104687.

\bibitem{ref26} Q. Xu, R. Zhang, Y. Zhang, Y. Wang, and Q. Tian, ``A Fourier-Based Framework for Domain Generalization,'' in Proc. IEEE/CVF Conf. Comput. Vis. Pattern Recognit. (CVPR), 2021, pp. 14383--14392, doi: 10.1109/CVPR46437.2021.01415.

\bibitem{ref27} S. Zhang, L. Zhang, and Z.-Y. Liu, ``Frequency-Based Pseudo-Domain Generation for Domain Generalizable Object Detection,'' Neurocomputing, vol. 542, Art. no. 126265, 2023, doi: 10.1016/j.neucom.2023.126265.

\bibitem{ref28} X. Xu, J. Yang, W. Shi, S. Ding, L. Luo, and J. Liu, ``PhysAug: A Physical-Guided and Frequency-Based Data Augmentation for Single-Domain Generalized Object Detection,'' in Proc. AAAI Conf. Artif. Intell., vol. 39, no. 20, pp. 21815--21823, 2025, doi: 10.1609/aaai.v39i20.35487.

\bibitem{ref29} X. Wang, K. Bin, X. Wang, P. Wan, T. Hu, and P. Zhong, ``Adversarial Spectral Perturbation for Single-Domain Generalized Object Detection,'' Pattern Recognit., vol. 171, pt. B, Art. no. 112264, 2026, doi: 10.1016/j.patcog.2025.112264.

\bibitem{ref30} O. Sim\'{e}oni et al., ``DINOv3,'' arXiv:2508.10104, 2025, doi: 10.48550/arXiv.2508.10104.

\bibitem{ref31} B. He, Y. Ji, Q. Ye, Z. Tan, and L. Wu, ``Generalized Diffusion Detector: Mining Robust Features from Diffusion Models for Domain-Generalized Detection,'' in Proc. IEEE/CVF Conf. Comput. Vis. Pattern Recognit. (CVPR), 2025, pp. 9921--9932, doi: 10.1109/CVPR52734.2025.00927.

\bibitem{ref32} M. Hong, F. Liu, C. Gevaert, G. Vosselman, and H. Cheng, ``Bridge: Basis-Driven Causal Inference Marries VFMs for Domain Generalization,'' in Proc. IEEE/CVF Conf. Comput. Vis. Pattern Recognit. (CVPR), 2026, pp. 31962--31973.

\bibitem{ref33} Y. Zhang, R. Han, N. Guo, W. Feng, S. Wang, and L. Wan, ``VFM\textsuperscript{4}SDG: Unveiling the Power of VFMs for Single-Domain Generalized Object Detection,'' arXiv:2604.21502, 2026, doi: 10.48550/arXiv.2604.21502.

\bibitem{ref34} F. Liu, C. Gao, Y. Zhang, J. Guo, J. Wang, and D. Meng, ``InfMAE: A Foundation Model in the Infrared Modality,'' in Proc. Eur. Conf. Comput. Vis. (ECCV), 2024, pp. 420--437, doi: 10.1007/978-3-031-72649-1\_24.

\bibitem{ref35} Y. Xing, X. Su, S. Zhang, D. Chu, and D. Xu, ``DuGI-MAE: Improving Infrared Mask Autoencoders via Dual-Domain Guidance,'' in Proc. AAAI Conf. Artif. Intell., vol. 40, no. 32, pp. 27135--27143, 2026, doi: 10.1609/aaai.v40i32.39928.

\bibitem{ref36} M. Zhang, Y. Wang, J. Guo, Y. Li, X. Gao, and J. Zhang, ``IRSAM: Advancing Segment Anything Model for Infrared Small Target Detection,'' in Proc. Eur. Conf. Comput. Vis. (ECCV), 2024, pp. 233--249, doi: 10.1007/978-3-031-72855-6\_14.

\bibitem{ref37} R. Fan, W. Zhao, M. Lin, Q. Wang, Y.-J. Liu, and W. Wang, ``Generalizable Thermal-Based Depth Estimation via Pre-Trained Visual Foundation Model,'' in Proc. IEEE Int. Conf. Robot. Autom. (ICRA), 2024, pp. 14614--14621, doi: 10.1109/ICRA57147.2024.10610394.

\bibitem{ref38} Y. Nalcakan, H. Ju, I. Park, S. Yeo, Y. Jin, and S. Kim, ``SpectraDINO: Modality-Conditioned Adaptation of RGB Vision Foundation Models Across Infrared Bands,'' arXiv:2605.02258, 2026, doi: 10.48550/arXiv.2605.02258.

\bibitem{ref39} Z. Chen et al., ``Vision Transformer Adapter for Dense Predictions,'' in Proc. Int. Conf. Learn. Represent. (ICLR), 2023. [Online]. Available: \url{https://arxiv.org/abs/2205.08534}

\bibitem{ref40} E. J. Hu et al., ``LoRA: Low-Rank Adaptation of Large Language Models,'' in Proc. Int. Conf. Learn. Represent. (ICLR), 2022. [Online]. Available: \url{https://arxiv.org/abs/2106.09685}

\bibitem{ref41} T.-Y. Lin, P. Doll\'{a}r, R. Girshick, K. He, B. Hariharan, and S. Belongie, ``Feature Pyramid Networks for Object Detection,'' in Proc. IEEE Conf. Comput. Vis. Pattern Recognit. (CVPR), 2017, pp. 2117--2125, doi: 10.1109/CVPR.2017.106.

\bibitem{ref42} Z. Nie, L. Xue, Z. Fang, J. Ren, Y. Wei, and J. Zheng, ``M3OT: A Multi-Drone Multi-Modality Dataset for Multi-Object Tracking,'' Sci. Data, vol. 12, no. 1, Art. no. 1927, Dec. 2025, doi: 10.1038/s41597-025-06204-0.

\bibitem{ref43} M. Everingham, L. van Gool, C. K. I. Williams, J. Winn, and A. Zisserman, ``The PASCAL Visual Object Classes (VOC) Challenge,'' Int. J. Comput. Vis., vol. 88, no. 2, pp. 303--338, Jun. 2010, doi: 10.1007/s11263-009-0275-4.

\bibitem{ref44} R. Padilla, W. L. Passos, T. L. B. Dias, S. L. Netto, and E. A. B. da Silva, ``A Comparative Analysis of Object Detection Metrics with a Companion Open-Source Toolkit,'' Electronics, vol. 10, no. 3, Art. no. 279, Jan. 2021, doi: 10.3390/electronics10030279.

\bibitem{ref45} X. Yu et al., ``Object Localization Under Single Coarse Point Supervision,'' in Proc. IEEE/CVF Conf. Comput. Vis. Pattern Recognit. (CVPR), 2022, pp. 4868--4877, doi: 10.1109/CVPR52688.2022.00482.

\bibitem{ref46} S. Ren, K. He, R. Girshick, and J. Sun, ``Faster R-CNN: Towards Real-Time Object Detection with Region Proposal Networks,'' in Adv. Neural Inf. Process. Syst., vol. 28, pp. 91--99, 2015.

\bibitem{ref47} Z. Rao, J. Guo, L. Tang, Y. Huang, X. Ding, and S. Guo, ``SRCD: Semantic Reasoning With Compound Domains for Single-Domain Generalized Object Detection,'' IEEE Trans. Neural Netw. Learn. Syst., vol. 36, no. 7, pp. 12497--12506, Jul. 2025, doi: 10.1109/TNNLS.2024.3480120.

\bibitem{ref48} Y. Fu, S. Wang, F. Wu, J. Lyu, Z. Liu, and M. K. Ng, ``Rethinking Representations for Cross-Domain Infrared Small Target Detection: A Generalizable Perspective from the Frequency Domain,'' arXiv:2604.01934, 2026, doi: 10.48550/arXiv.2604.01934.

\bibitem{ref49} L. Liu, S. Wang, and Y. Fu, ``Hierarchical Structure Dependency Whitening for Single-Domain Generalized Infrared Small Target Detection,'' IEEE Signal Process. Lett., vol. 33, pp. 396--400, 2026, doi: 10.1109/LSP.2025.3647431.

\bibitem{ref50} K. Li, F. Han, Y. Wang, K. Huang, B. Yang, and Y. Tian, ``MMIDNet: A Multi-level Mutual Information Disentanglement Network for Cross-domain Infrared Small Target Detection,'' IEEE J. Sel. Topics Appl. Earth Observ. Remote Sens., vol. 18, pp. 26309--26323, 2025, doi: 10.1109/JSTARS.2025.3620484.

\bibitem{ref51} W. Chi, J. Liu, X. Lu, Y. Ni, and X. Wang, ``A Contrast-Invariant Feature Extraction Framework for Single-Domain Generalization in Infrared Small Target Detection,'' Knowl.-Based Syst., vol. 310, Art. no. 112963, Feb. 2025, doi: 10.1016/j.knosys.2025.112963.

\end{thebibliography}
\end{document}